\documentclass[journal]{IEEEtran}
\usepackage{graphicx} 
\usepackage{color}
\usepackage{colortbl} 
\usepackage{dcolumn}
\usepackage[table]{xcolor}
\usepackage{booktabs}
\usepackage{multirow}
\usepackage{pifont}
\usepackage[export]{adjustbox}
\usepackage{epstopdf}
\usepackage{subfigure}
\usepackage{amsfonts}  
\usepackage{caption}
\usepackage{amsmath}
\usepackage{algorithmic}
\usepackage{array}
\usepackage{stfloats}
\usepackage{url}

\usepackage{tabularx,booktabs}
\usepackage[linesnumbered,lined,ruled]{algorithm2e}
\usepackage{threeparttable}
\usepackage{rotating}
\usepackage{footmisc}
\usepackage{float}
\usepackage{makecell}
\usepackage[switch]{lineno}
\usepackage{algorithmic}
\usepackage{bm}

\makeatletter 
\newcommand{\removelatexerror}{\let\@latex@error\@gobble}
\makeatother

\usepackage{verbatim}

\begin{document}
\captionsetup[figure]{labelformat={default},labelsep=period,name={Fig.},labelfont={small},  singlelinecheck=off, textfont={small}} 
\captionsetup[table]{format=plain,labelformat=simple,justification=centering, labelsep=newline, singlelinecheck=false,labelfont={small}, textfont={sc, small}}

\title{CoMiX: Cross-Modal Fusion with Deformable Convolutions for HSI-X Semantic Segmentation}

\author{Xuming~Zhang,~\IEEEmembership{}
        Xingfa~Gu,~\IEEEmembership{} 
        Qingjiu~Tian,~\IEEEmembership{}
        and Lorenzo Bruzzone,~\IEEEmembership{}
            
\thanks{
Xuming Zhang and Qingjiu Tian are with the International Institute for Earth System Science, Nanjing University, Nanjing 210023, China (e-mail: xuming.zhang.rs@gmail.com).

Xingfa Gu is with the Key Laboratory of Computational Optical Imaging Technology, Aerospace Information Research Institute, Chinese Academy of Sciences, Beijing 100094, China. 

Lorenzo Bruzzone is with the Department of Information Engineering and Computer Science, University of Trento, 38050 Trento, Italy. 
}}

\markboth{JOURNAL OF LATEX CLASS FILES}%
{Shell \MakeLowercase{\textit{et al.}}: Bare Demo of IEEEtran.cls for Journals}

\maketitle
\begin{abstract}
Pixel-wise semantic segmentation of hyperspectral images (HSIs) enables detailed and accurate analysis of complex scenes, benefiting various applications. Improving HSI semantic segmentation by exploiting complementary information from a supplementary data type (here denoted as X-modality) is promising but challenging due to differences in imaging sensors, image content, and resolution. Current cross-model fusion techniques have limitations in enhancing modality-specific and modality-shared information, as well as in capturing dynamic information interactions and fusion between different modalities. In response, this study proposes CoMiX, a cross-modal fusion framework with deformable convolutions (DCNs) for pixel-wise HSI-X semantic segmentation. CoMiX is an asymmetric encoder-decoder architecture designed to extract, calibrate, and fuse information from HSI and X modalities. Its pipeline includes an encoder with two parallel and interacting backbones and a lightweight all-multilayer perceptron (ALL-MLP) decoder. Within the encoder, four stages are deployed, each incorporating 2D DCN blocks for the X-model to adapt to geometric variations, and 3D DCN blocks for HSIs to adaptively aggregate spatial-spectral features. Additionally, each stage includes a Cross-Modality Feature enhancement and eXchange (CMFeX) module and a feature fusion module (FFM). CMFeX is designed to exploit spatial and spectral correlations from both modalities to recalibrate and enhance modality-specific and modality-shared features while adaptively exchanging complementary information between them. Outputs from CMFeX are fed into the FFM for fusion and passed to the next stage for further information learning. Finally, the outputs from each FFM are integrated by the ALL-MLP decoder for final prediction. Extensive experiments demonstrate that our CoMiX achieves superior performance and generalizes well to various multimodal recognition tasks. The CoMiX code will be released.
\end{abstract}

\IEEEpeerreviewmaketitle
\section{Introduction}
Hyperspectral images (HSIs), with continuous spectral information over a wide range of wavelengths, enable detailed analysis and discrimination of different materials \cite{my3}. HSI semantic segmentation is a fundamental task in scene understanding and supports many applications \cite{9783460, 9885024}, including land-cover analysis \cite{my1}, agricultural quality assessment \cite{WIEME2022156}, and urban development monitoring \cite{r3}. It is closely related to HSI classification as both tasks predict the category at the pixel-level rather than at the image-level \cite{rSegFormer}. However, using single HSI data has reached its performance bottleneck, especially in complex scenes where classes exhibit similar spectral signatures. Multimodal data fusion techniques address these challenges by integrating information from a supplementary sensor type (here called X-modality), such as digital surface models (DSM), light detection and ranging (LiDAR), and synthetic aperture radar (SAR), to provide complementary information to HSIs. For example, DSM offers elevation information, LiDAR provides accurate 3D spatial information, and SAR captures structural information about the Earth's surface \cite{rS2FL}.

Numerous multimodal fusion approaches \cite{9594516, 10008228} have been developed, typically classified into three main types: pixel-level fusion \cite{LI2017100}, feature-level fusion \cite{7902153}, and decision-level fusion \cite{618255}.  Pixel-level fusion combines data by averaging or concatenating along the input channels, often requiring alignment and registration of images from different modalities. Although effective in some scenarios, pixel-level fusion faces challenges in handling differences in image resolution and characteristics. Feature-level fusion employs individual encoders for each modality to capture the related features, which are then fused to enhance information representation. Decision-level fusion combines results generated independently by each modality to make a final decision. It is often used in scenarios where individual modalities excel in different aspects of a task, and their combined decisions lead to improved performance. Among them, feature-level fusion has emerged as one of the most promising approach due to its scalability and effectiveness in integrating information from multiple modalities. For example,  Rasti \textit{et al.} \cite {7902153} adopted extinction profiles (EPs) \cite {7730335} to automatically extract spatial and elevation information from HSI and LiDAR data, respectively, which was subsequently fused to generate classification maps. The shared and specific feature learning model, S2FL \cite{rS2FL}, gathers information from HSI and another modality by decomposing modality-shared components on the latent manifold subspace while separating modality-specific components. Furthermore, many follow-up studies \cite{CHEN201727, 8000656} have progressively introduced more sophisticated strategies \cite{rManifoldLearning, 7010915} to improve the integration of information across different modalities. Despite offering valuable solutions, these traditional fusion techniques still face inherent drawbacks, such as the requirement for expert feature engineering and limitations in feature representation and generalization.

Deep learning (DL), an automatic feature learning technique, has demonstrated overwhelming superiority in vision tasks \cite {10231003, 9772757, 9862940, 9405447, 9526853}. Among DL algorithms, convolutional neural networks (CNNs) are preferred for their scalability, flexibility and ease of optimization \cite{rCCT}. Numerous CNN-based studies have been developed for multi-source remote sensing image learning and fusion. Chen \textit{et al.} \cite {7940007} employed two-branch CNNs to extract features from multispectral image (MSI)/HSI and LiDAR data, which are fused through fully connected (FC) layers. An FC-dominated encoder–decoder network called EndNet \cite{9179756} extracts information from different modalities and combines them by forcing the fused features to reconstruct the multimodal input. Ge \textit{et al.} \cite{9336235} extracted EPs and local binary patterns (LBPs) from HSI and LiDAR data and fused them using a deep residual fusion framework. The unified multimodal fusion framework presented in \cite{9174822}, developed as a baseline for complex scenes, includes extraction and fusion subnetworks, with the latter incorporating a novel cross-fusion strategy that outperforms concatenation-based fusion. In contrast to these approaches, Fusion-FCN, the winner of the 2018 IEEE Data Fusion Contest (DFC), was built with a fully convolutional network (FCN) [18], demonstrating the significant potential of FCNs in multimodal semantic segmentation. Despite these advances, CNNs treat input content uniformly, which is impractical given the different contributions of inputs to identification \cite{my2}.

Several efforts \cite{rSENet, rDANet} have been made to address the equal treatment of convolution kernels by introducing attention mechanisms to guide models on “where" and “what" to focus. Wang \textit{et al.} \cite {WANG2023385} and Zhang \textit{et al.} \cite {9813366} applied attention modules to enhance optical and LiDAR feature learning, respectively. Coupled adversarial learning-based classification (CALC) \cite{rCALC} introduced a spatial attention module for adversarial feature learning to capture high-order semantic information from HSI and LiDAR data. Some studies \cite{9150738, 9915611, WANG20221} developed cross-modal attention modules for multimodal information fusion. Xue \textit{et al.} \cite{9526908} introduced self-attention modules to simultaneously enhance feature learning and fusion of HSI and LiDAR data. Roy \textit{et al.} \cite{10462184} developed a cross hyperspectral and LiDAR attention, where LiDAR patch tokens serve as queries and HSI patch tokens serve as keys and values.

Taking advantage of the adaptive spatial aggregation and long-range dependencies inherent in self-attention, the vision transformer (ViT) \cite {rViT} was developed and further refined and scaled in subsequent studies \cite{rSwin, 9578646}. With the advent of transformers, multimodal data fusion has also witnessed a significant progress. The multimodal fusion transformer (MFT) \cite{rMFT} and the deep hierarchical vision transformer \cite {9755059} were developed to learn heterogeneous information from HSI and LiDAR data, followed by a cross-attention module for heterogeneous feature fusion. Similarly, the local information interaction transformer network \cite{10004002}, a dual-branch transformer, exploits sequence information from multimodal data, which is then fused and fed into a convolutional transformer module for classification. 

Although transformers have shown impressive results in multimodal information representation, their low sensitivity to local features and lack of the inductive biases inherent in CNNs limit their performance. Therefore, some hybrid networks \cite {9150738, 9926173, 10445018} have been developed to combine the strengths of CNNs and transformers to overcome these limitations. For instance, the local–global transformer network, i.e., GLT-Net, \cite{9926173} deploys multiscale convolution blocks in shallow layers to learn local spatial features, incorporates transformer modules in deep layers to capture global spectral features, and then performs global–local information fusion for prediction. Similarly, Fusion\_HCT \cite {9999457} sequentially employs CNN and transformer blocks to learn HSI and LiDAR information in parallel, followed by a cross-token attention fusion module for information fusion. 

Alternatively, some studies \cite {9880273, rUniRepLKNet} in computer vision have attempted to introduce long-range dependencies into CNNs by using convolutions with large dense kernels (e.g., $31 \times 31$). However, these large kernel CNNs are  still inferior in performance to state-of-the-art transformers \cite {rDCNv3}. Recently, Mamba \cite{gu2023mamba} has been introduced as an efficient alternative to transformers for encoding sequential features in a linear fashion. Nonetheless, the inherent 1-D nature of its selective scanning technique poses challenges when applied to 2D or higher-dimensional visual data, potentially resulting in the loss of crucial spatial information. Furthermore, the deformable convolution network (DCN) series \cite{rDCNv3, rDCNv2, rDCNv4} innovatively combines the strengths of CNNs and transformers. DCNs achieve adaptive spatial aggregation by incorporating learnable offsets into convolutional kernels. These offsets allow the network to dynamically adjust and learn appropriate receptive fields (local- or long-range) according to the input data. Additionally, the convolutional kernels used in DCN remain small, typically $3 \times 3$, avoiding the optimization challenges and high computational cost \cite{rDCNv3}.

Although the aforementioned studies have achieved improvements in feature extraction or multimodal data fusion by combining the advantages of CNNs and transformers from different perspectives, they still have important limitations:

1) Most of the aforementioned approaches are  predominantly designed for specific modalities, such as HSI and LiDAR data fusion. However, a general and modality-agnostic multimodal architecture for HSI-X semantic segmentation is highly desirable to facilitate robust real-world scene understanding. Such an architecture streamlines research efforts by eliminating the need to optimize for specific modality combinations, while allowing seamless integration of new sensors as they emerge. 

2) Existing multimodal semantic segmentation methods often process HSIs similarly to 2D X-modal, overlooking that HSIs are 3D spatial-spectral data. This limits the comprehensive use of the rich spectral information inherent in HSIs, thus compromising the integration of discriminative information. Therefore, HSI-X semantic segmentation architectures should not only accommodate various X modalities but also effectively exploit the abundant spectral information in HSIs, while maintaining computational efficiency for processing 3D data.

3) Although significant progress has been achieved in HSI-X information fusion, directly concatenating or adding the features from two modalities  \cite {9179756, 9813366, rCALC, 9926173, rFusionFCN} or simply assembling information from two modalities through a cross-modality fusion module \cite {9150738, 10004002, rFusionHCT} may lead to inferior performance. Hence, it is important to design a cross-modality fusion technique that can  effectively identify modality-specific, modality-shared, and complementary features, and integrate them into an efficient representation.

\begin{figure*}[htbp] 
\centerline{\includegraphics[width=18cm]{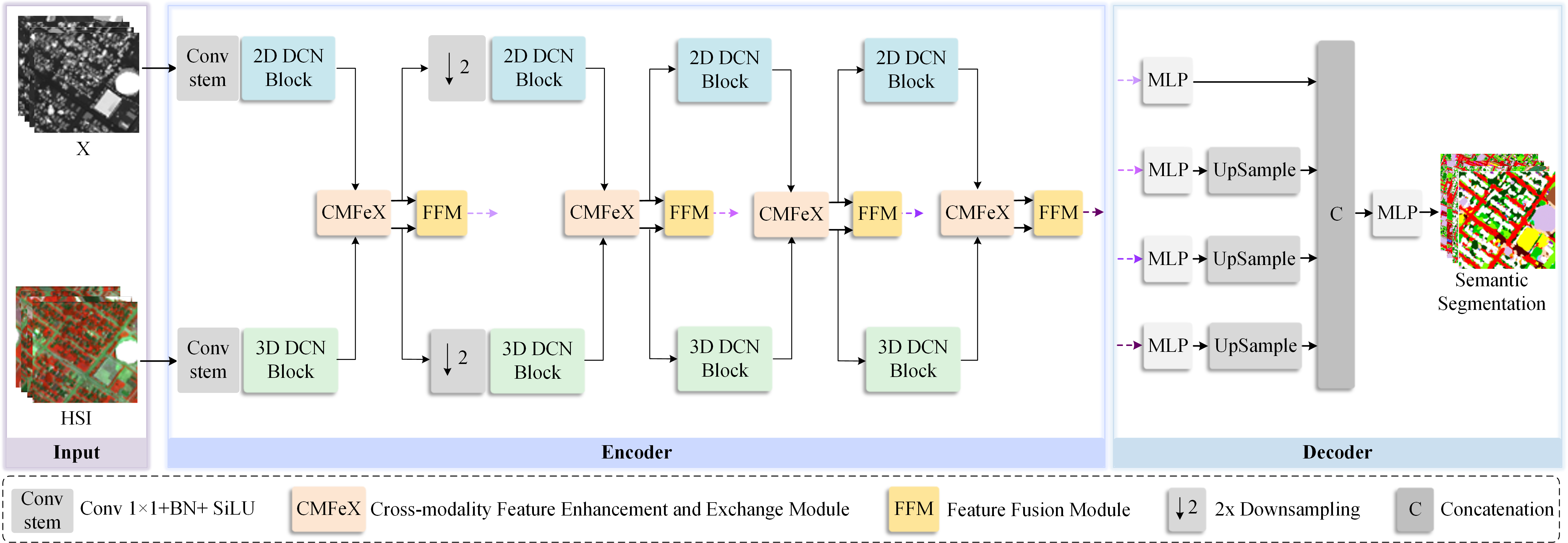}} 
\caption{Overview of the proposed CoMiX framework for HSI-X semantic segmentation.}
\label{Overview} 
\end{figure*}

In this work, we explore an alternative paradigm for designing a universal modality-agnostic HSI-X semantic segmentation framework, called CoMiX. The backbone of CoMiX incorporates deformable convolutions with custom block-level designs inspired by transformers. This integration facilitates the development of 3D DCN blocks and 2D DCN blocks for adaptive extraction of modality-specific features from HSI and X data. As illustrated in Fig.~\ref{Overview}, CoMiX consists of an encoder with two parallel and interactive backbones, and a lightweight all-multilayer perceptron (ALL-MLP) decoder. Within the encoder, the 3D DCN blocks and 2D DCN blocks are tailored to adaptively extract modality-specific features from HSI and X data, respectively.  Additionally, a Cross-Modality Feature enhancement and eXchange (CMFeX) module is introduced at each stage to recalibrate modality-specific feature responses and promote cross-modality feature interaction across spatial and spectral dimensions. The enhanced HSI- and X-modality features are then passed to the feature fusion module (FFM) for cross-modality information fusion, while being fed to the next stage for further information learning. Finally, the All-MLP decoder aggregates the fused features from each stage for prediction. 

The main contributions of this study are as follows.

1) Proposing a universal modality-agnostic HSI-X semantic segmentation framework, CoMiX, that is specifically tailored to extract modality-specific, modality-shared, and complementary features from HSI and X data while advancing the cross-modality information fusion.

2) Developing two parallel and interactive backbones with deformable convolution, where 3D DCN blocks and 2D DCN blocks are designed for modality-specific feature learning from HSI and X data, respectively.

3) Designing the CMFeX module to recalibrate and highlight modality-specific and modality-shared information, while facilitating the exchange of complementary information between them in a more effective and efficient manner. The combination of CMFeX and FFM further enhances cross-modality information fusion.

The remainder of this study is organized as follows.  Section \uppercase\expandafter{\romannumeral2} details the proposed CoMiX framework. Section \uppercase\expandafter{\romannumeral3} presents the experimental setup and results. Section \uppercase\expandafter{\romannumeral4} focuses on the network analysis. Finally, concluding remarks and future work are presented in Section \uppercase\expandafter{\romannumeral5}.

\section{Methodology}

Fig.~\ref{Overview} presents an overview of the proposed CoMiX framework, an asymmetric encoder-decoder structure. The encoder employs two parallel and interacting backbones to adaptively aggregate local- and long-range dependencies, as well as fine and coarse features from HSI- and X-modality (e.g., LiDAR, DEM, SAR data). The decoder employs the lightweight All-MLP \cite{rSegFormer} to integrate multi-level features for predicting semantic segmentation masks. Within the encoder, one convolution stem and four stages are deployed. The convolution stem, which includes a $1 \times 1$ convolution, batch normalization, and a SiLU activation function, projects HSI and X data onto the same channels. In each stage, the proposed 2D DCN and 3D DCN blocks are used to perform feature learning from X and HSI data, respectively. Instead of gradually decreasing the resolution of the feature maps in each stage, we only reduce the resolution by a factor of two before the second stage to achieve a better trade-off between performance and computational complexity. Additionally, at each stage, we introduce the CMFeX module, which is specifically designed to exploit spatial and spectral correlations from both modalities to recalibrate and enhance modality-specific and modality-shared features while facilitating the exchange of complementary information between them. The output of the CMFeX is sent to the FFM for cross-modality information fusion and simultaneously fed to the next stage for further information learning. Finally, the fused information generated by each FFM is forwarded to the decoder for pixel-level prediction. Details of the 2D DCN block, 3D DCN block, CMFeX, FFM, and the lightweight All-MLP decoder are given in the subsequent sections.

\subsection{2D DCN Block} 
Geometric variations induced by scale, viewpoint, etc., pose significant challenges in land-cover recognition. Currently, DCNs are the state-of-the-art method for addressing this problem \cite{rDCNv2}. The DCN series extends traditional convolutional operations by introducing learnable offsets that dynamically adjust sampling locations within inputs. This allows the network to perform adaptive spatial aggregation and effectively adapt to geometric variations. DCNv1 \cite{rDCNv1} is the pioneering work that introduced the deformable convolution. Subsequently, DCNv2 \cite{rDCNv2}, DCNv3 \cite{rDCNv3}, and DCNv4 \cite{rDCNv4} incorporated tailored modifications to further improve information representation and bridge the gap between CNNs and ViTs. Inspired by \cite{rDCNv4}, we refine it to develop a faster and more effective feature extraction operator for the X modality.

\begin{figure*}[htbp] 
\centerline{\includegraphics[width=18cm]{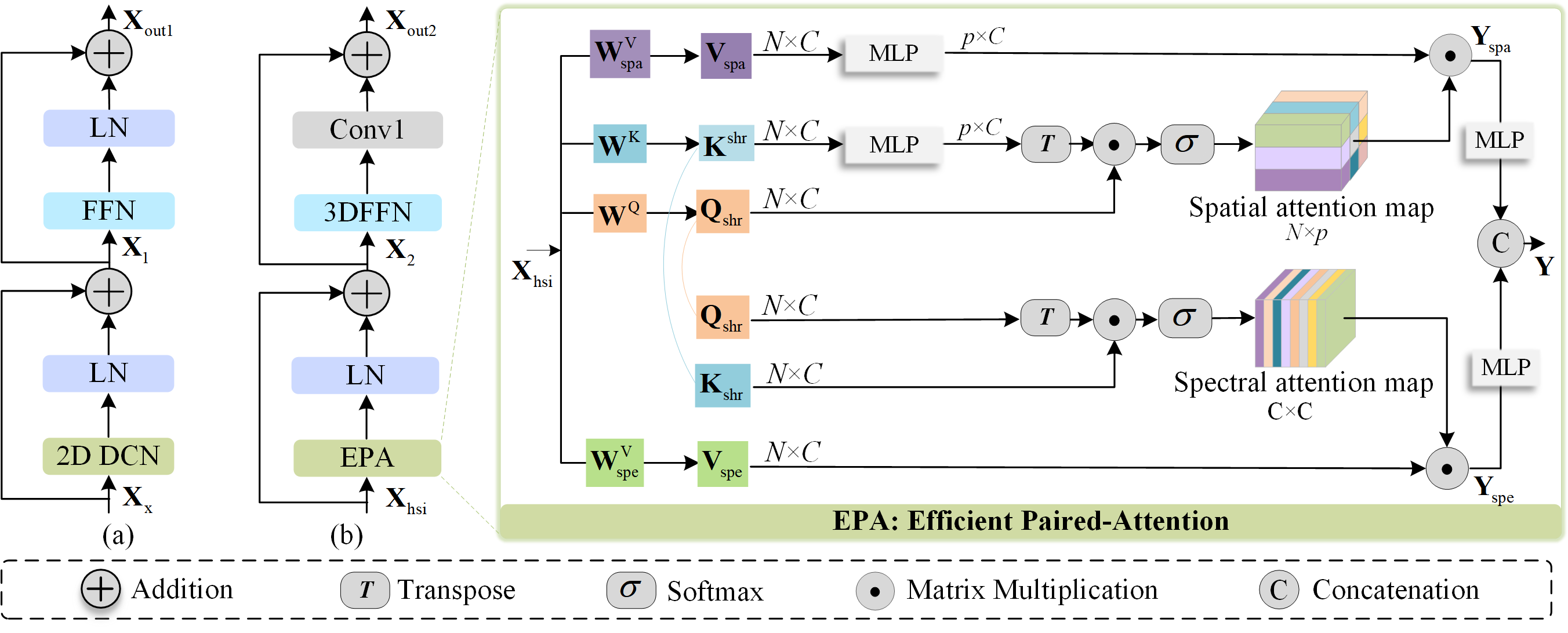}} 
\caption{Illustration of the (a) 2D DCN block and (b) 3D DCN block.}
\label{3DDCN} 
\end{figure*}

Given a convolution kernel with $K$ sampling locations, let ${{p}_{k}}$ denote the pre-specified offset for the $k$-th location. For example, a regular $3 \times 3$ convolution kernel is defined as $K = 9$ and ${{p}_{k}}\in$ {(-1, -1), (-1, 0), . . ., (1, 1)}. Let $x\left( p \right)$ and $y\left( p \right)$ denote the features at location $p$ of the input $\mathbf{x}$ and output $\mathbf{y}$, respectively. Following the multi-group designed in group convolution, the DCNv4 splits the spatial aggregation process into multi-groups to learn richer information. Then, DCNv4 at location $p$ can be expressed as:
\begin{equation}
y\left( p \right)=\sum\limits_{g=1}^{G}{\sum\limits_{k=1}^{K}{{{w}_{g}}}}{{m}_{gk}}{{x}_{g}}\left( p+{{p}_{k}}+\Delta {{p}_{gk}} \right),
\end{equation}
where $G$  represents the number of spatial aggregation groups. For the $g$-th group, ${{w}_{g}}\in {{\mathbb{R}}^{C\times {{C}^{'}}}}$ is the location-irrelevant projection weight, ${{m}_{gk}}\in \mathbb{R}$ is the modulation scalar for the $k$-th sampling point, and ${{x}_{g}}\in {{\mathbb{R}}^{{{C}^{'}}\times H\times W}}$ denotes the input feature maps, where ${{C}^{'}}=C/G$ represents the group dimension. $\Delta {{p}_{gk}}$ signifies the sampling offset corresponding to the pre-specified offset ${{p}_{k}}$ within the $g$-th group. In practice, the modulation scalar ${{m}_{gk}}$ and the sampling offset $\Delta {{p}_{gk}}$ are generated from the input $\mathbf{x}$ by a $3\times 3$ depth-wise convolution (DWConv) followed by a linear projection.

Equation (1) illustrates that the sampling offset $\Delta {{p}_{gk}}$ is flexible and can capture both short- and long-range dependencies. Additionally, the sampling offset $\Delta {{p}_{gk}}$ and the modulation scalar ${{m}_{gk}}$ are input-dependent and learnable, allowing the network to effectively model geometric transformations. Meanwhile, DCNv4 retains the inductive bias of convolution, leading to improved efficiency with reduced training time and data requirements. It is also easier to optimize and more computationally and memory-efficient than other techniques, such as self-attention \cite{rViT}, large kernel convolution \cite{9880273}, and Mamba \cite{gu2023mamba, liu2024vmamba}.

By combining the aforementioned DCNv4 with advanced transformer block designs, the 2D DCN block can be expressed as:
\begin{equation}
{{\mathbf{X}}_{1}}=\text{LN(DCNv4(}{{\mathbf{X}}_{\text{x}}}\text{))}+{{\mathbf{X}}_{\text{x}}},
\end{equation}
\begin{equation}
{{\mathbf{X}}_{\text{out1}}}=\text{LN(FFN(}{{\mathbf{X}}_{1}}\text{))}+{{\mathbf{X}}_{1}},
\end{equation}
\begin{equation}
\text{FFN}=\text{MLP}\left( \text{GELU}\left( \text{MLP}\left( {{\mathbf{X}}_{1}} \right) \right) \right),
\end{equation}
where ${{\mathbf{X}}_{\text{x}}}$, ${{\mathbf{X}}_{1}}$ and ${{\mathbf{X}}_{\text{out1}}}$ represent the input X-modality data, intermediate features, and output features of the 2D DCN block, respectively. GELU is an activation function and LN denotes layer normalization \cite{rLN}. We stack 2D DCN blocks in each stage for X-modality feature learning.

Unlike previous DCN series using sparse operators with $3\times 3$ windows throughout the entire network, we employ $3\times 3$ windows in the first two stages and $7\times 7$ windows in the last two stages. In this way, the network can selectively focus on more important regions and capture finer details, enhancing information learning. This is particularly important for downstream tasks like HSI semantic segmentation, where scenes exhibit complex structures and contain smaller objects that require context-aware feature extraction and precise localization. 

\begin{figure*}[htbp] 
\centerline{\includegraphics[width=18cm]{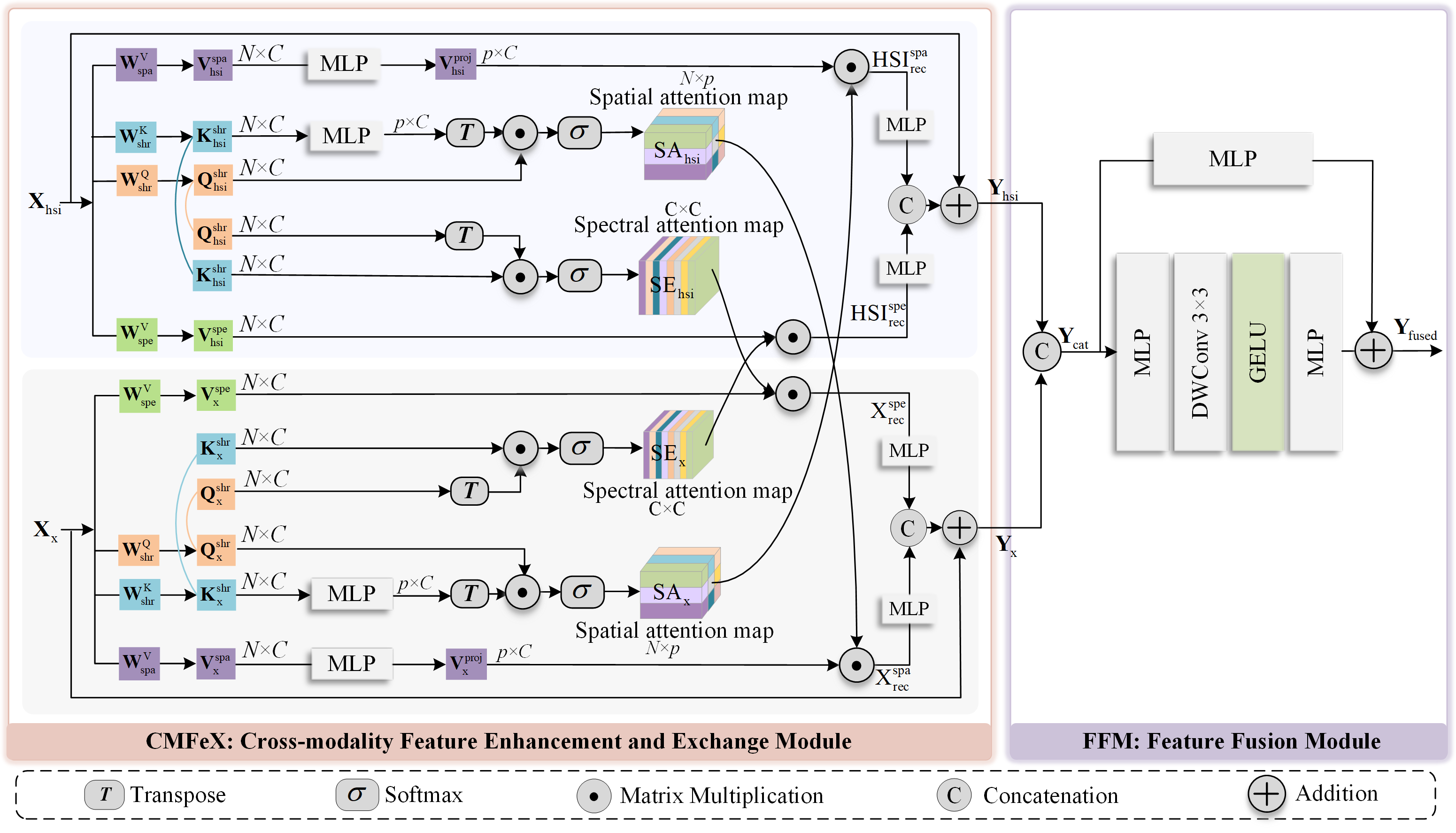}} 
\caption{Illustration of the proposed CMFeX and FFM modules.}
\label{CMFeX} 
\end{figure*}

\subsection{3D DCN Block} 
Given that HSI data is a 3D cube, it is intuitive to extend the 2D DCN block to 3D for HSI feature extraction. However, this extension significantly increases the number of parameters and the computational cost as the number of channels scales along the third dimension \cite {rDLKA}. To solve this problem, we introduce the alternative efficient paired-attention (EPA)  module \cite{10526382} for HSI processing. EPA adopts a shared key-query strategy between spatial and spectral attention modules, optimizing efficiency in terms of parameters, FLOPs, and inference speed. 

As illustrated in Fig.~\ref{3DDCN}, EPA consists of a spatial attention module and a spectral attention module, both of which use shared keys and shared queries to minimize computational overhead and capture complementary features. Given an input ${{\mathbf{X}}_{\text{hsi}}}\in {{\mathbb{R}}^{HWD\times C}}$, we linearly transform it into spatial values ${{\mathbf{V}}_{\text{spa}}}\in {{\mathbb{R}}^{N\times C}}$, shared queries ${{\mathbf{Q}}_{\text{shr}}}\in {{\mathbb{R}}^{N\times C}}$, shared keys ${{\mathbf{K}}_{\text{shr}}}\in {{\mathbb{R}}^{N\times C}}$ and spectral values ${{\mathbf{V}}_{\text{spe}}}\in {{\mathbb{R}}^{N\times C}}$ via ${{\mathbf{V}}_{\text{spa}}}=\mathbf{W}_{\text{spa}}^{\text{V}}{\mathbf{X}}_{\text{hsi}}$, ${{\mathbf{Q}}_{\text{shr}}}={{\mathbf{W}}^{\text{Q}}}{\mathbf{X}}_{\text{hsi}}$, ${{\mathbf{K}}_{\text{shr}}}={{\mathbf{W}}^{\text{K}}}{\mathbf{X}}_{\text{hsi}}$ and ${{\mathbf{V}}_{\text{spe}}}=\mathbf{W}_{\text{spe}}^{\text{V}}{\mathbf{X}}_{\text{hsi}}$, where $N=HWD$, $\mathbf{W}_{\text{spa}}^{\text{V}}$, ${{\mathbf{W}}^{\text{Q}}}$, ${{\mathbf{W}}^{\text{K}}}$, and $\mathbf{W}_{\text{spe}}^{\text{V}}$ are the corresponding projection weights. For simplicity, we omit the transformation from $H \times W \times D \times C$ to $N \times C$. ${{\mathbf{V}}_{\text{spa}}}$ and ${{\mathbf{V}}_{\text{spe}}}$  are specific to the spatial and spectral attention modules, respectively, while ${{\mathbf{Q}}_{\text{shr}}}$ and ${{\mathbf{K}}_{\text{shr}}}$ are shared between them.  
The computational complexity of regular self-attention is $O\left({N^{2}}\right)$, indicating that the computational complexity and memory requirements increase quadratically with the input resolution. Consequently, it quickly becomes a computational bottleneck for high-resolution inputs. To mitigate this, ${{\mathbf{K}}_{\text{shr}}}$, ${{\mathbf{V}}_{\text{spa}}}$ and ${{\mathbf{V}}_{\text{spe}}}$ are projected onto lower dimensions before the attention operation, allowing the aggregation of long-range dependencies within a controlled computational budget.  

For the spatial attention, ${{\mathbf{K}}_{\text{shr}}}$ and ${{\mathbf{V}}_{\text{spa}}}$ are projected from $N\times C$ onto a low-dimensional matrix of shape $p\times C$ using MLP. Then, the spatial self-attention module is estimated as:
\begin{equation}
{{\mathbf{Y}}_{\text{spa}}}=\text{Softmax}\left( \frac{1}{\sqrt{d}}{{\mathbf{Q}}_{\text{shr}}}\mathbf{K}{{_{\text{shr}}^{\text{proj}}}^{\text{T}}} \right)\mathbf{V}_{\text{spa}}^{\text{proj}},
\end{equation}
where ${{\mathbf{Y}}_{\text{spa}}}$ represents the output of the spatial attention module, $d$ provides a normalization, $\mathbf{V}_{\text{spa}}^{\text{proj}}=\text{MLP}\left( {{\mathbf{V}}_{\text{spa}}} \right)$ and $\mathbf{K}_{\text{shr}}^{\text{proj}}=\text{MLP}\left( {{\mathbf{K}}_{\text{shr}}} \right)$ denote the projected values and projected shared keys, respectively. 

Similarly, the spectral attention module captures the interdependencies between channels. By using the same ${{\mathbf{Q}}_{\text{shr}}}$ and ${{\mathbf{K}}_{\text{shr}}}$ as the spatial attention module, the channel attention is computed as:
\begin{equation}
{{\mathbf{Y}}_{\text{spe}}}={{\mathbf{V}}_{\text{spe}}}\cdot \text{Softmax}\left( \frac{1}{\sqrt{d}}\mathbf{Q}_{\text{shr}}^{\text{T}}{{\mathbf{K}}_{\text{shr}}} \right),
\end{equation}
where ${{\mathbf{Y}}_{\text{spe}}}$ represents the output of the spectral attention module. 

Finally, the MLP is employed to transform the outputs of the two attention modules, halving the channel dimension and obtaining enriched feature representations. A concatenation operation is then performed to fuse them. Therefore, the final output $\mathbf{Y}$ of the EPA module is obtained by:
\begin{equation}
\mathbf{Y}=\left[ \text{MLP}\left( {{\mathbf{Y}}_{\text{spa}}} \right),\text{MLP}\left( {{\mathbf{Y}}_{\text{spe}}} \right) \right],
\end{equation}
where $\left[ \cdot ,\cdot  \right]$ denotes the concatenation operation along the channel dimension. 

With the aforementioned EPA module, the 3D DCN block can be formulated as:
\begin{equation}
{{\mathbf{X}}_{2}}=\text{LN(EPA(}{{\mathbf{X}}_{\text{hsi}}}\text{))}+{{\mathbf{X}}_{\text{hsi}}},
\end{equation}
\begin{equation}
{{\mathbf{X}}_{\text{out2}}}=\text{Con}{{\text{v}}_{1}}\text{(3DFFN(}{{\mathbf{X}}_{2}}\text{))}+{{\mathbf{X}}_{2}},
\end{equation}
\begin{equation}
\text{3DFFN}=\text{Con}{{\text{v}}_{3}}\left( \text{ReLU}\left( \text{Con}{{\text{v}}_{3}}\left( {{\mathbf{X}}_{2}} \right) \right) \right),
\end{equation}
where ${{\mathbf{X}}_{2}}$ and ${{\mathbf{X}}_{\text{out2}}}$ denote the intermediate features and outputs of the 3D DCN block, respectively. The $\text{Con}{{\text{v}}_{1}}$ and $\text{Con}{{\text{v}}_{2}}$ represent $1\times 1\times 1$ and  $3\times 3\times 3$ convolutions, respectively.

\subsection{Cross-modality Feature Enhancement and Exchange (CMFeX) Module} 
Information from different modalities provides unique and complementary insights into the observed scene. Nevertheless, each modality may introduce noise and other undesirable elements, such as blur, distortion, and occlusion. Integrating information from multiple modalities can effectively enhance the quality and richness of information while filtering out noise and mitigating undesired information. Thus, multimodal data fusion is essential for extracting comprehensive and reliable information from complex scenes. Consequently, we develop the CMFeX module to recalibrate and enhance modality-specific features while promoting interaction between different modalities. 

As shown in Fig.~\ref{CMFeX}, CMFeX processes HSI and X modalities in spatial and spectral dimensions for feature calibration and interaction. Given an HSI input ${{\mathbf{X}}_{\text{hsi}}}\in {{\mathbb{R}}^{HWD\times C}}$, it is linearly transformed into spatial values $\mathbf{V}_{\text{hsi}}^{\text{spa}}\in {{\mathbb{R}}^{N\times C}}$,  shared queries $\mathbf{Q}_{\text{hsi}}^{\text{shr}}\in {{\mathbb{R}}^{N\times C}}$, shared keys $\mathbf{K}_{\text{hsi}}^{\text{shr}}\in {{\mathbb{R}}^{N\times C}}$ and spectral values $\mathbf{V}_{\text{hsi}}^{\text{spe}}\in {{\mathbb{R}}^{N\times C}}$. Similar to EPA, before performing the attention operation, $\mathbf{K}_{\text{hsi}}^{\text{shr}}$ and $\mathbf{V}_{\text{hsi}}^{\text{spa}}$ are projected onto lower-dimensional matrices of shape $p\times C$ via an MLP to reduce computational complexity. Then, the spatial and spectral attention maps of HSI can be obtained by:
\begin{equation}
\text{S}{{\text{A}}_{\text{hsi}}}=\text{Softmax}\left( \frac{1}{\sqrt{d}}\mathbf{Q}_{\text{hsi}}^{\text{shr}}\mathbf{K}{{_{\text{hsi}}^{\text{proj}}}^{\text{T}}} \right),
\end{equation}
\begin{equation}
\text{S}{{\text{E}}_{\text{hsi}}}=\text{Softmax}\left( \frac{1}{\sqrt{d}}\mathbf{Q}{{_{\text{hsi}}^{\text{shr}}}^{\text{T}}}\mathbf{K}_{\text{hsi}}^{\text{shr}} \right),
\end{equation}
where $\text{S}{{\text{A}}_{\text{hsi}}}$ and $\text{S}{{\text{E}}_{\text{hsi}}}$ represent the spatial and spectral attention maps of HSIs, respectively, and $\mathbf{K}_{\text{hsi}}^{\text{proj}}=\text{MLP}\left( \mathbf{K}_{\text{hsi}}^{\text{shr}} \right)$ denotes the projected shared keys of HSIs.

Similarly, the spatial values $\mathbf{V}_{\text{x}}^{\text{spa}}\in {{\mathbb{R}}^{HW\times C}}$,  shared queries $\mathbf{Q}_{\text{x}}^{\text{shr}}\in {{\mathbb{R}}^{HW\times C}}$, shared keys $\mathbf{K}_{\text{x}}^{\text{shr}}\in {{\mathbb{R}}^{HW\times C}}$, and spectral values $\mathbf{V}_{\text{x}}^{\text{spe}}\in {{\mathbb{R}}^{HW\times C}}$ of the X modality can be obtained by linearly transforming the input ${{\mathbf{X}}_{\text{x}}}\in {{\mathbb{R}}^{H\times W\times C}}$.  The spatial and spectral attention maps for the X modality can also be obtained by:
\begin{equation}
\text{S}{{\text{A}}_{\text{x}}}=\text{Softmax}\left( \frac{1}{\sqrt{d}}\mathbf{Q}_{\text{x}}^{\text{shr}}\mathbf{K}{{_{\text{x}}^{\text{proj}}}^{\text{T}}} \right),
\end{equation}
\begin{equation}
\text{S}{{\text{E}}_{\text{x}}}=\text{Softmax}\left( \frac{1}{\sqrt{d}}\mathbf{Q}{{_{\text{x}}^{\text{shr}}}^{\text{T}}}\mathbf{K}_{\text{x}}^{\text{shr}} \right),
\end{equation}
where $\text{S}{{\text{A}}_{\text{x}}}$ and $\text{S}{{\text{E}}_{\text{x}}}$ denote the spatial and spectral attention maps of the X modality, respectively, and $\mathbf{K}_{\text{x}}^{\text{proj}}=\text{MLP}\left( \mathbf{K}_{\text{x}}^{\text{shr}} \right)$ represents the projected shared keys of the X modality.

Spatial rectification is achieved through a spatial cross-attention process, where the spatial values are multiplied by the spatial attention map of the other modality. This process can be represented as:
\begin{equation}
\text{HSI}_{\text{rec}}^{\text{spa}}={{\text{SA}}_{\text{x}}} \cdot \mathbf{V}_{\text{hsi}}^{\text{proj}},
\end{equation}
\begin{equation}
\text{X}_{\text{rec}}^{\text{spa}}={{\text{SA}}_{\text{hsi}}}\cdot\mathbf{V}_{\text{x}}^{\text{proj}} ,
\end{equation}
where $\text{HSI}_{\text{rec}}^{\text{spa}}$ and $\text{X}_{\text{rec}}^{\text{spa}}$ represent the recalibrated spatial information of HSI and X, respectively, $\mathbf{V}_{\text{hsi}}^{\text{proj}}=\text{MLP}\left( \mathbf{V}_{\text{hsi}}^{\text{spa}} \right)$ and $\mathbf{V}_{\text{x}}^{\text{proj}}=\text{MLP}\left( \mathbf{V}_{\text{x}}^{\text{spa}} \right)$ denote the projected  spatial values of HSI and X, respectively.

Similar to the spatial rectification, the spectral rectification is formulated as:
\begin{equation}
\text{HSI}_{\text{rec}}^{\text{spe}}=\mathbf{V}_{\text{hsi}}^{\text{spe}}\cdot\text{S}{{\text{E}}_{\text{x}}},
\end{equation}
\begin{equation}
\text{X}_{\text{rec}}^{\text{spe}}=\mathbf{V}_{\text{x}}^{\text{spe}}\cdot\text{S}{{\text{E}}_{\text{hsi}}},
\end{equation}
where $\text{HSI}_{\text{rec}}^{\text{spe}}$ and $\text{X}_{\text{rec}}^{\text{spe}}$ represent the recalibrated spectral information of HSI and X, respectively.

The recalibrated HSI and X are obtained by transforming them using an MLP and then performing a concatenation operation to fuse them: 
\begin{equation}
\mathbf{X}_{\text{hsi}}^{\text{rec}}=\left[ \text{MLP}\left( \text{HSI}_{\text{rec}}^{\text{spa}} \right), \text{MLP}\left( \text{HSI}_{\text{rec}}^{\text{spe}} \right) \right],
\end{equation}
\begin{equation}
\mathbf{X}_{\text{x}}^{\text{rec}}=\left[ \text{MLP}\left( \text{X}_{\text{rec}}^{\text{spa}} \right), \text{MLP}\left( \text{X}_{\text{rec}}^{\text{spe}} \right) \right].
\end{equation}

Finally, the rectification features $\mathbf{X}_{\text{hsi}}^{\text{rec}}$ and $\mathbf{X}_{\text{x}}^{\text{rec}}$  perform a residual connection with the inputs ${{\mathbf{X}}_{\text{hsi}}}$ and ${{\mathbf{X}}_{\text{x}}}$, respectively, to facilitate gradient flow and address the vanishing gradient problem, i.e.,
\begin{equation}
{{\mathbf{Y}}_{\text{hsi}}}=\mathbf{X}_{\text{hsi}}^{\text{rec}}+{{\mathbf{X}}_{\text{hsi}}},
\end{equation}
\begin{equation}
{{\mathbf{Y}}_{\text{x}}}=\mathbf{X}_{\text{x}}^{\text{rec}}+{{\mathbf{X}}_{\text{x}}}, 
\end{equation}
where ${{\mathbf{Y}}_{\text{hsi}}}$ and ${{\mathbf{Y}}_{\text{x}}}$ are the final outputs of the CMFeX. ${{\mathbf{Y}}_{\text{hsi}}}$ and ${{\mathbf{Y}}_{\text{x}}}$ are then sent to the next stage for futher information learning, while being fed to FFM for cross-modality information fusion.

\subsection{Feature Fusion Module (FFM)}
After obtaining the outputs ${{\mathbf{Y}}_{\text{hsi}}}$ and ${{\mathbf{Y}}_{\text{x}}}$ of CMFeX, we concatenate them along the channel dimension. Then, a simple depthwise feedforward network (DWFFN) with residual connection is used to produce the fused cross-modality features. The DWFFN is realized by two MLP layers separated by a $3 \times 3$ DWConv and a GELU activation function. Therefore, FFM is formulated as: 
\begin{equation}
{{\mathbf{Y}}_{\text{fused}}}=\text{ DWFFN}\left({{\mathbf{Y}}_{\text{cat}}} \right)+\text{MLP}\left({{\mathbf{Y}}_{\text{cat}}} \right),
\end{equation}
\begin{equation}
{{\mathbf{Y}}_{\text{cat}}}=\left[ \mathbf{Y}{_{\text{hsi}}}\text{,}{{\mathbf{Y}}_{\text{x}}} \right],
\end{equation}
\begin{equation}
\text{DWFFN}=\text{MLP}\left( \text{DWConv}\left( \text{GELU}\left( \text{MLP}\left({{\mathbf{Y}}_{\text{cat}}}\right) \right) \right) \right),
\end{equation}
where ${{\mathbf{Y}}_{\text{fused}}}$ is the output of FFM. In this way, the concatenated features ${{\mathbf{Y}}_{\text{cat}}}$ of size ${\mathbb{R}}^{H\times W\times 2C}$ are fused into the final output ${{\mathbf{Y}}_{\text{fused}}}$ of size  ${\mathbb{R}}^{H\times W\times C}$ for further feature decoding.

\subsection{Lightweight All-MLP Decoder}
Given that our encoder allows for a larger effective receptive field, we adopt the lightweight ALL-MLP decoder \cite{rSegFormer} for predicting the segmentation results. As illustrated in Fig.~\ref{Overview}, the decoder contains only a few MLP layers. The multi-level fused features from each stage are first fed into an MLP layer to unify their channel dimension and then upsampled. The results are concatenated and then passed through another two successive MLP layers for fusion and prediction, respectively. As evident, the decoder avoids the use of computationally intensive components, allowing our CoMiX to achieve lower computational cost, fewer parameters, and increased efficiency.  Furthermore, this streamlined approach facilitates seamless integration into various applications.

\section{EXPERIMENTS AND RESULTS}
This section details the configuration of our CoMiX framework for HSI-X semantic segmentation. Under this configuration, experiments were conducted and the corresponding quantitative and qualitative results are presented.

\begin{table}[th]
 \centering
  \caption{The land-cover types and the number of training and test samples in the Houston2013 dataset}
 \begin{tabular}{ccccc}
 \toprule[1pt]
    ID    & Color & Land-cover Type & Training & Test \\ \midrule[0.5pt]
    C1    & \cellcolor[rgb]{ 0,  .8,  0} & Healthy grass & 198   & 1053 \\
    C2    & \cellcolor[rgb]{ .498,  1,  0} & Stressed grass & 190   & 1064 \\
    C3    & \cellcolor[rgb]{ .18,  .541,  .341} & Artificial turf & 192   & 505 \\
    C4    & \cellcolor[rgb]{ 0,  .275,  0} & Evergreen trees & 188   & 1056 \\
    C5    & \cellcolor[rgb]{ .624,  .322,  .176} & Deciduous trees & 186   & 1056 \\
    C6    & \cellcolor[rgb]{ 0,  1,  1} & Bare earth & 182   & 143 \\
    C7    &                             & Water & 196   & 1072 \\
    C8    & \cellcolor[rgb]{ .851,  .745,  .922} & Residential buildings & 191   & 1053 \\
    C9    & \cellcolor[rgb]{ 1,  0,  0} & Non-residential buildings & 193   & 1059 \\
    C10   & \cellcolor[rgb]{ .314,  0,  0} & Roads & 191   & 1036 \\
    C11   & \cellcolor[rgb]{ 1,  .847,  .694} & Sidewalks & 181   & 1054 \\
    C12   & \cellcolor[rgb]{ 1,  1,  0} & Crosswalks & 192   & 1041 \\
    C13   & \cellcolor[rgb]{ .929,  .6,  0} & Major thoroughfares & 184   & 285 \\
    C14   & \cellcolor[rgb]{ .569,  .118,  .706} & Highways & 181   & 247 \\
    C15   & \cellcolor[rgb]{ 0,  .361,  1} & Railways & 187   & 473 \\   \midrule[0.5pt]
    \multicolumn{3}{c}{Total}  & 28332 & 12197 \\
\bottomrule[1pt]
\end{tabular}
\label{Houston_Tab}
\end{table}

\begin{figure}[h] 
\subfigure
{
\begin{minipage}[hb]{1\linewidth}
\flushleft
\includegraphics[width=8.5cm]{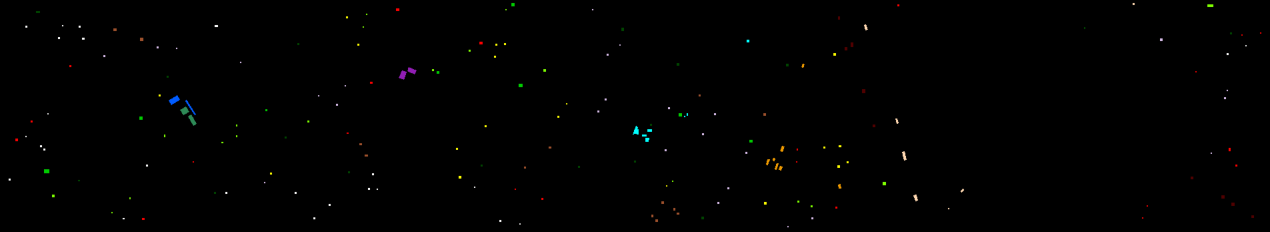}
\end{minipage}
}
\subfigure
{
\begin{minipage}[b]{1\linewidth}
\flushleft
\includegraphics[width=8.5cm]{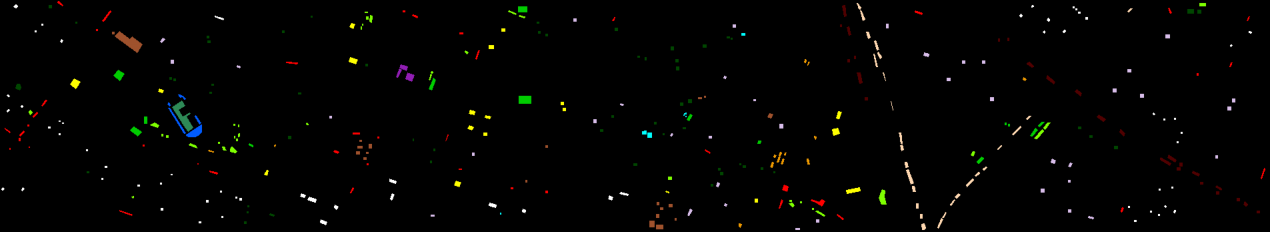}
\end{minipage}
}
\caption{Houston2013 dataset: spatial distribution of (a) the training samples and (b) the test samples.}
\label{Houston_Trn} 
\end{figure}

\subsection{Description of Datasets} 
To assess the effectiveness of our CoMiX framework, we conducted experiments on three public multi-modality benchmarks that combine HSI-DSM, HSI-SAR, and HSI-multispectral LiDAR (MS-LiDAR) data, respectively.

The Houston2013 dataset was captured by the National Center for Airborne Laser Mapping in June 2012 and covers the University of Houston campus and its adjacent urban region. This dataset comprises two distinct data sources: an HSI and a DSM derived from LiDAR, both sharing identical spatial size ($349 \times 1905$ ) and spatial resolution (2.5 m). The HSI contains 144 spectral bands, spanning the wavelength range from 380 nm to 1050 nm. This scene contains 15 distinct classes, as shown in Fig.~\ref{Houston_Trn}. More specific details are provided in Table~\ref{Houston_Tab}.

\begin{table}[th] 
 \centering
 \caption{The land-cover types and the number of training and test samples in the Berlin dataset}
 \begin{tabular}{ccccc}
 \toprule[1pt]
    ID    & Color & Land-cover Type & Training & Test \\ \midrule[0.5pt]
    C1    & \cellcolor[rgb]{ 0,  .4,  0} & Forest & 443   & 54511 \\
    C2    & \cellcolor[rgb]{ .89,  .89,  .89} & Residential Area & 423   & 268219 \\
    C3    & \cellcolor[rgb]{ .549,  .263,  .18} & Industrial Area & 499   & 19067 \\
    C4    & \cellcolor[rgb]{ .451,  1,  .675} & Low Plants & 376   & 58906 \\
    C5    & \cellcolor[rgb]{ 1,  1,  .49} & Soil  & 331   & 17095 \\
    C6    & \cellcolor[rgb]{ .235,  .353,  .447} & Allotment & 280   & 13025 \\
    C7    & \cellcolor[rgb]{ .733,  .082,  .906} & Commercial Area & 298   & 24526 \\
    C8    & \cellcolor[rgb]{ 0,  .361,  1} & Water & 170   & 6502 \\ \midrule[0.5pt]
    \multicolumn{3}{c}{Total}  & 2820 & 461851 \\
\bottomrule[1pt]
\end{tabular}
\label{Berlin_Tab}
\end{table}

\begin{figure}[h] 
\subfigure
{
\begin{minipage}[hb]{1\linewidth}
\centering
\includegraphics[width=8cm]{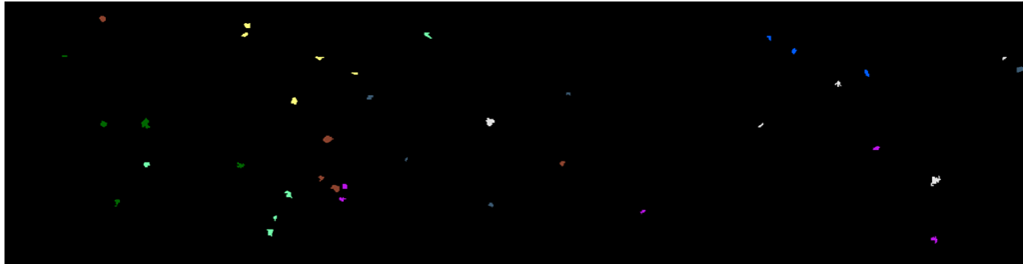}
\end{minipage}
}
\subfigure
{
\begin{minipage}[b]{1\linewidth}
\centering
\includegraphics[width=8cm]{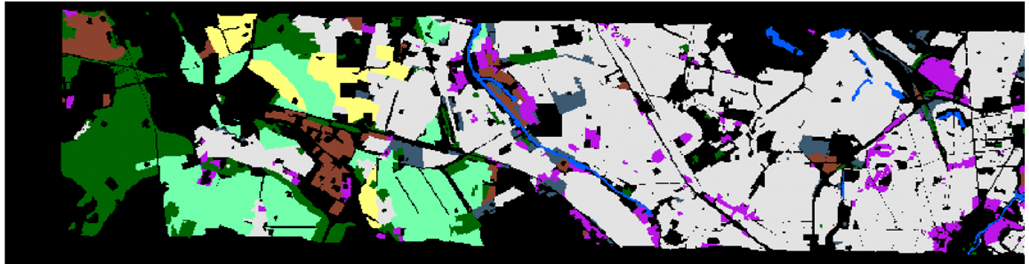}
\end{minipage}
}
\caption{Berlin dataset: spatial distribution of (a) the training samples and (b) the test samples.}
\label{Berlin_trn} 
\end{figure}

The Berlin dataset, obtained from the urban and surrounding rural areas of Berlin, is a comprehensive collection of HSI and SAR images. The HSI was generated by synthesizing EnMAP data simulated from HyMap HSI data. Meanwhile, the SAR image was derived from Sentinel-1 satellite dual-polarization (VV-VH) single-view complex data. The HSI is a $797 \times 220\times 224$ data cube with a spatial resolution of 30 meters and covers a wavelength range of 0.4 to 2.5 µm. In contrast, the SAR image consists of $1723\times 476$ pixels with a spatial resolution of 13.89 m. To match the spatial resolution of the SAR image, the HSI was interpolated using the nearest neighbor technique. Extensive multi-step data preprocessing, including orbital profiling, radiometric calibration, denoising, speckle reduction and topographic recalibration, was performed to ensure the quality and reliability of the dataset for subsequent analyses. Table~\ref{Berlin_Tab} provides detailed information on land-cover classes and their corresponding number of samples. Furthermore, the spatial distribution and visual representation of these classes are shown in Fig.~\ref{Berlin_trn}.

\begin{table}[ht] 
 \centering
 \caption{The land-cover types and the number of training and test samples in the DFC2018 dataset}
 \begin{tabular}{ccccc}
 \toprule[1pt]
    ID    & Color & Land-cover Type & Training & Test \\ \midrule[0.5pt]   
    C1    & \cellcolor[rgb]{ 0,  .8,  0} & Healthy grass & 39196 & 20000 \\
    C2    & \cellcolor[rgb]{ .498,  1,  0} & Stressed grass & 130008 & 20000 \\
    C3    & \cellcolor[rgb]{ .18,  .541,  .341} & Artificial turf & 2736  & 20000 \\
    C4    & \cellcolor[rgb]{ 0,  .541,  0} & Evergreen trees & 54322 & 20000 \\
    C5    & \cellcolor[rgb]{ 0,  .275,  0} & Deciduous trees & 20172 & 20000 \\
    C6   & \cellcolor[rgb]{ .624,  .322,  .176} & Bare earth & 18064 & 20000 \\
    C7    & \cellcolor[rgb]{ 0,  1,  1} & Water & 1064  & 1628 \\
    C8    &       & Residential buildings & 158995 & 20000 \\
    C9    & \cellcolor[rgb]{ .851,  .745,  .922} & Non-residential buildings & 894769 & 20000 \\
    C10   & \cellcolor[rgb]{ 1,  0,  0} & Roads & 183283 & 20000 \\
    C11   & \cellcolor[rgb]{ .663,  .624,  .584} & Sidewalks & 136035 & 20000 \\
    C12   & \cellcolor[rgb]{ .498,  .498,  .498} & Crosswalks & 6059  & 5345 \\
    C13   & \cellcolor[rgb]{ .624,  0,  0} & Major thoroughfares & 185438 & 20000 \\
    C14   & \cellcolor[rgb]{ .314,  0,  0} & Highways & 39438 & 20000 \\
    C15   & \cellcolor[rgb]{ 1,  .847,  .694} & Railways & 27748 & 11232 \\
    C16   & \cellcolor[rgb]{ 1,  1,  0} & Paved parking lots & 45932 & 20000 \\
    C17   & \cellcolor[rgb]{ .929,  .6,  0} & Unpaved parking lots & 587   & 3524 \\
    C18   & \cellcolor[rgb]{ 1,  0,  1} & Cars  & 26289 & 20000 \\
    C19   & \cellcolor[rgb]{ 0,  0,  1} & Trains & 21479 & 20000 \\
    C20   & \cellcolor[rgb]{ .686,  .765,  .867} & Stadium seats & 27296 & 20000 \\  \midrule[0.5pt]
    \multicolumn{3}{c}{Total}  & 2018910 & 341729 \\
\bottomrule[1pt]
\end{tabular}
\label{DFC_Tab}
\end{table}

\begin{figure}[ht] 
\centering
\includegraphics[width=8.5cm]{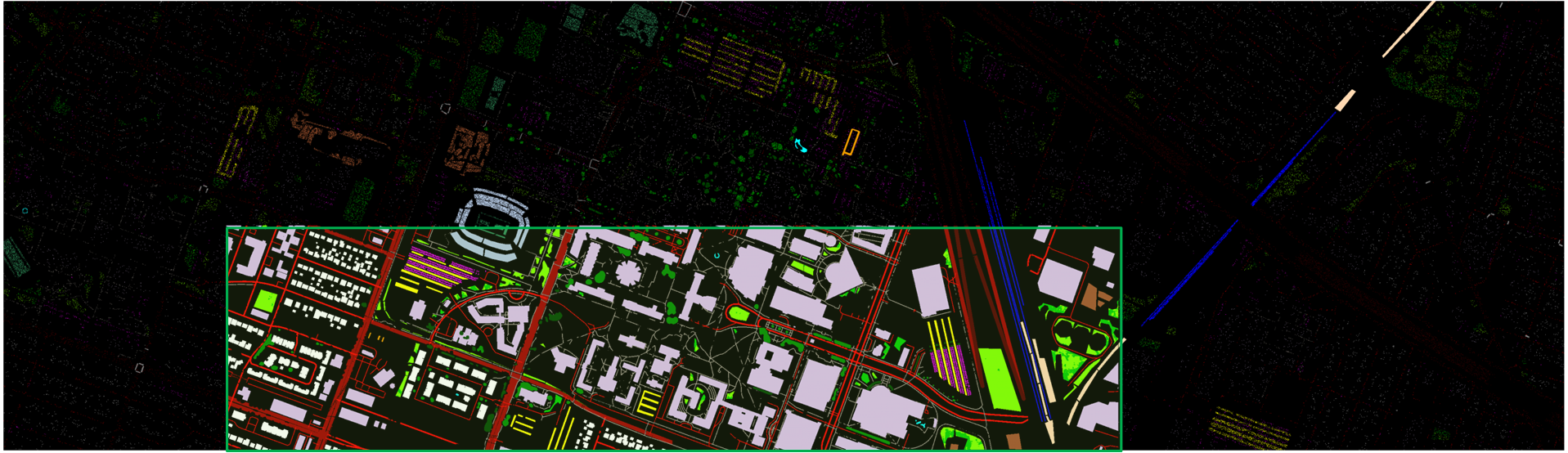}
\caption{DFC2018 dataset: spatial distribution of training (green box) and test samples (outside the green box).}
\label{DFC_trn} 
\end{figure}

The DFC2018 dataset, generated by the National Center for Airborne Laser Mapping in February 2017, covers the University of Houston campus and its adjacent urban areas. The dataset adopted in this study includes HSI data with a spatial resolution of 1 m and MS-LiDAR point cloud data with a spatial resolution of 0.5 m. The HSI dataset encompasses $4172\times 1202$ pixels distributed across 48 bands, covering the spectral range from 380 to 1050 nm. Meanwhile, MS-LiDAR contains a DSM, a digital elevation model (DEM), and three intensity rasters at distinct wavelengths: 1550 nm (near-infrared), 1064 nm (mid-infrared), and 532 nm (green). To match the spatial resolution of MS-LiDAR images, the HSI was interpolated to a spatial resolution of 0.5 m using the nearest neighbor interpolation algorithm. To obtain the actual elevation of objects, the normalized DSM (NDSM) value was calculated via $\text{NDSM}=\text{DSM}-\text{DEM}$. This dataset includes 20 land-cover classes as summarized in Table~\ref{DFC_Tab}. Furthermore, the spatial distribution of training and test data is visually depicted in Fig.~\ref{DFC_trn}.

Before the experiments, these three datasets were normalized to [0, 1] to standardize the magnitude of the data, thereby enhancing network convergence during training.

\subsection{Experimental Settings} 

\begin{table*}[tbp]
  \centering
  \caption{Classification accuracy produced by different methods on the Houston2013 dataset. The numbers within ( ) are the standard deviations of the corresponding metrics. The best result in each row is marked in bold}
   \begin{tabular}{p{4.19em}cccccccccccc}
    \toprule[1pt]
    \makecell{Metrics \\ and Class} & \makecell{SVM-X \\ \cite{rSVM}}  & \makecell{SVM-HSI \\ \cite{rSVM}} & \makecell{SVM \\ \cite{rSVM}} & \makecell{FusAtNet \\ \cite{9150738}} & \makecell{CACL \\ \cite{rCALC}} & \makecell{Fusion\_HCT \\ \cite{rFusionHCT}}  & \makecell{MFT \\ \cite{rMFT}}  & \makecell{Fusion-FCN \\ \cite{rFusionFCN}} & \makecell{Flex-MCFNet\\ \cite{10216780}} & \makecell{CoMiX \\ (Ours)} \\
    \midrule[0.5pt] 

     OA (\%)  & \makecell{17.16  \\ (0)} & \makecell{55.51 \\ (0)} & \makecell{64.68 \\ (0)} & \makecell{89.02\\ (1.03)} & \makecell{86.24 \\ (1.56)} & \makecell{88.95 \\ (1.48)} & \makecell{87.76 \\ (0.98)}  & \makecell{83.84 \\ (1.67)}  & \makecell{91.76 \\ (1.31)}  & \makecell{\textbf{95.75} \\ (1.18)} \\
    AA (\%)  & \makecell{16.66 \\ (0)} & \makecell{59.83 \\ (0)} & \makecell{67.14 \\ (0)} & \makecell{92.28\\ (1.18)} & \makecell{88.10 \\ (2.18)} & \makecell{90.48 \\ (1.91)} & \makecell{89.28 \\ (1.06)} &  \makecell{85.42 \\ (2.16)}  & \makecell{93.17 \\ (1.45)}  & \makecell{\textbf{96.23} \\ (1.06)} \\
    $ \kappa\times 100$ & \makecell{10.76 \\ (0)} & \makecell{52.12 \\ (0)} & \makecell{61.82 \\ (0)} & \makecell{88.09 \\ (1.14)} & \makecell{85.11 \\ (1.92)} & \makecell{88.00 \\ (1.62)} & \makecell{86.42 \\ (0.98)} & \makecell{82.50 \\ (1.61)}  & \makecell{91.11 \\ (1.56)}  & \makecell{\textbf{95.39} \\ (1.36)} \\
    \midrule[0.5pt]

    C1 & 55.56  & 81.96  & 82.43  & 82.96 & 80.63  & 82.48  & 82.67  & 83.00 & 85.77 & \textbf{90.60} \\
    C2 & 0.00  & 75.00   & 80.83 & \textbf{97.70} & 80.83  & 83.93  & 84.48  & 84.02 & 87.40 & 85.34 \\
    C3 & \textbf{100.00} & 99.01  & 99.01 & \textbf{100.00} & 85.55  & 98.22  & 98.68  & \textbf{100.00} & 99.86& \textbf{100.00} \\
    C4 & 15.63  & 89.21  & 88.35  & 95.74  & 91.00  & 92.36  & 92.33 & 90.72 & 94.37 & \textbf{100.00} \\
    C5 & 0.00  & 85.23  & 90.06   & 98.96  & 99.34  & 99.91  & 98.22  & 99.34 & 99.92 & \textbf{100.00} \\
    C6 & 0.00  & 78.32  & 78.32   & \textbf{100.00} & 93.01  & 91.72  & 95.45  & 99.30 & \textbf{100.00} & 98.60 \\
    C7 & 47.30  & 28.64  & 69.50  & 92.19 & 83.12  & 87.35  & 85.11  & 82.46 & 93.22& \textbf{93.75}  \\
    C8 & 31.43  & 12.92  & 54.42  & 82.81 & 85.28  & 91.19 & 78.73  & 90.98 & 93.81 & \textbf{93.92} \\
    C9 & 0.00  & 81.30  & 83.00   & 85.35 & 86.03  & 79.55  & 87.73 & 75.54 & 95.53 & \textbf{96.03}  \\
    C10 & 0.00  & 1.83  & 2.90    & 66.12  & 60.91  & 67.31  & 60.94  & 54.83 & 71.39 & \textbf{99.81} \\
    C11 & 0.00  & 55.88  & 64.52  & 85.06 & 94.88  & \textbf{96.40}  & 96.01 & 92.60 & 83.81 & 94.97 \\
    C12 & 0.00  & 0.10  & 0.00    & 89.63  & 88.38  & 95.89  & 92.88  & 70.70 & 96.76 & \textbf{99.23} \\
    C13 & 0.00  & 11.93  & 15.09  & 86.32 & 92.98 & 90.88  & 90.01  & 59.30 & \textbf{95.74} & 91.23 \\
    C14 & 0.00  & 97.17  & 99.60  & 99.19  & 99.60  & \textbf{100.00} & 99.87 & \textbf{100.00} & \textbf{100.00} & \textbf{100.00} \\
    C15 & 0.00  & 98.94  & 99.15  & 99.58 & \textbf{100.00} & \textbf{100.00} & 96.09  & 98.52 & \textbf{100.00} &  \textbf{100.00} \\
    \bottomrule[1pt]
    \end{tabular}%
  \label{Houston_acc}
\end{table*}%


\subsubsection{Comparison Methods}
We compared our CoMiX framework with the support vector machine (SVM) \cite{rSVM}, and several DL-based models, including FusAtNet \cite{9150738}, CALC \cite{rCALC}, Fusion\_HCT \cite{rFusionHCT}, MFT \cite{rMFT}, Fusion-FCN \cite{rFusionFCN}, and a  multistage information complementary fusion network based on flexible-mixup (Flex-MCFNet) \cite{10216780}. SVM is a pixel-wise classifier that works at the pixel level. FusAtNet, CALC, Fusion\_HCT, and MFT are patch-based classification networks designed specifically for HSI and LiDAR data fusion. More specifically, FusAtNet generates HSI-derived and LiDAR-derived attention maps that enhance the spectral and spatial information of HSIs, respectively. CACL develops a coupled adversarial feature learning subnetwork to extract semantic features from HSI and LiDAR data, followed by a multi-level feature fusion classification subnetwork. Fsion\_HCT sequentially employs CNN and transformer blocks for feature extraction, followed by a cross-token attention fusion module. MFT adopts a ViT-based network where the input patches come from HSIs and class tokens come from LiDAR data. Unlike the aforementioned networks, Fusion-FCN is an FCN-based architecture with consistent spatial sizes between input images and feature maps at each level. Finally, Flex-MCFNet provides comprehensive multimodal information fusion for HSI and X-modal (e.g., MSI, SAR, and LiDAR data) while still following the patch-based classification framework.

\subsubsection{Implementation Details}
We implemented our CoMiX framework on the PyTorch platform using the Adam optimizer and the cross-entropy loss function. CoMiX was trained for 500 epochs with a weight decay of 0.01. The initial learning rate was set to $6\times 10^{-5}$ for the Houston2013 and DFC2018 datasets, and $1\times 10^{-3}$ for the Berlin dataset. The learning rate was updated during training using the poly-learning rate schedule. The input size and batch size were set to $128\times 128$ and 4, respectively, due to GPU memory constraints. For the comparison methods, we used publicly available source codes and set their hyperparameters according to the relevant literature. To ensure fair comparisons, we applied the same preprocessing strategy and the same training and test sample sets. Additionally, we performed five independent tests for each experiment and reported the average metrics to ensure the objectivity of our results.

\subsubsection{Metric}
We measured the performance of different approaches using four main metrics: overall accuracy (OA), average accuracy (AA), kappa coefficient ($\kappa $), and producer accuracy for each category. 

\subsection{Comparison with Literature Methods} 
\subsubsection{Quantitative Results and Analysis}

\begin{figure*}[tb]      
\centerline{\includegraphics[width=17.5cm]{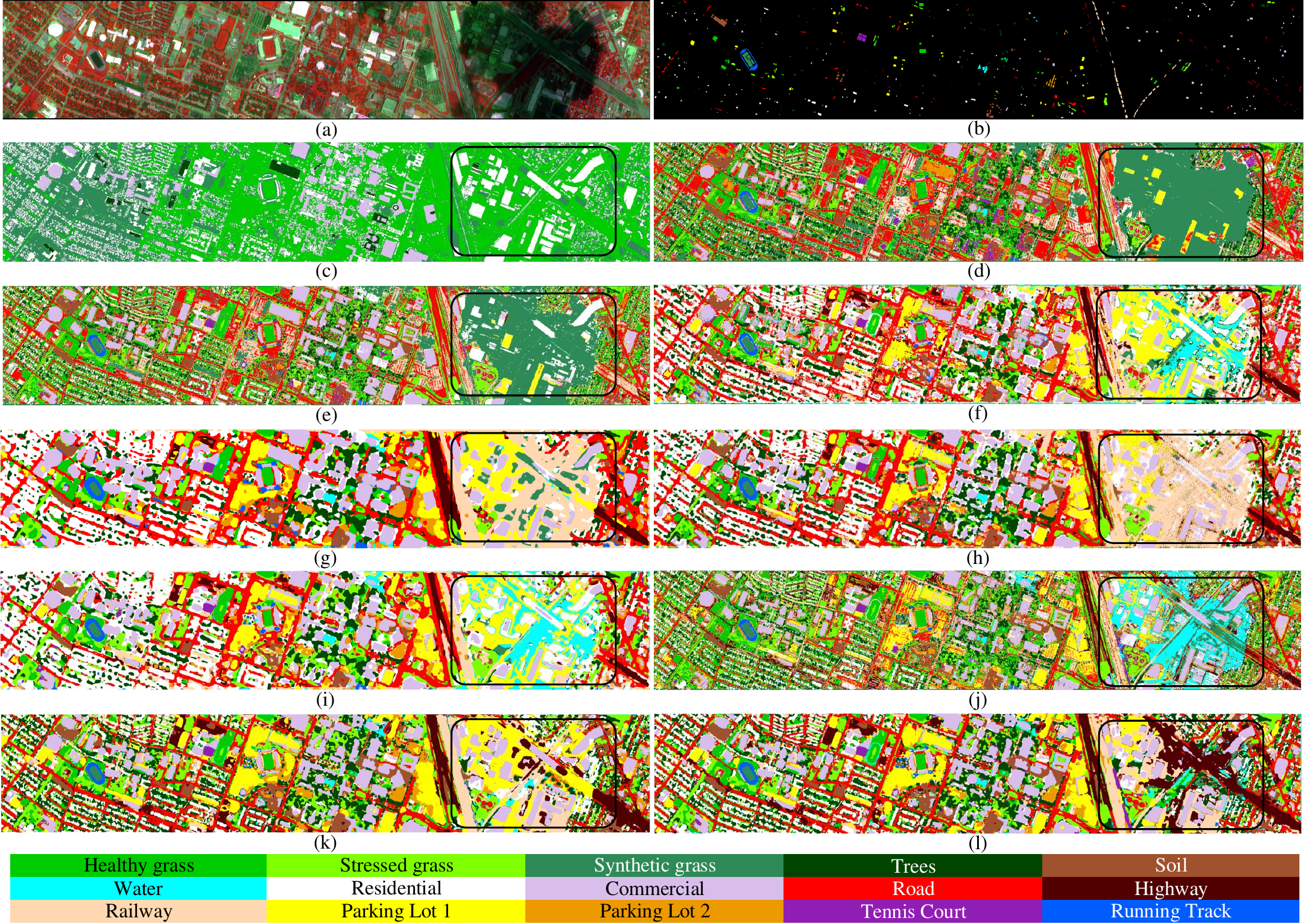}} 
\caption{Classification maps of different methods on the Houston2013 dataset: (a) False Color Image, (b) Ground Truth, (c) SVM-X, (d) SVM-HSI, (e) SVM, (f) FusAtNet, (g) CACL, (h) Fusion\_HCT, (i) MFT, (j) Fusion-FCN, (k) Flex-MCFNet, and (l) CoMiX.}
\label{Houston_map} 
\end{figure*}

The quantitative results of the comparative experiments are presented in Tables~\ref{Houston_acc}–\ref{DFC_acc}, with the best results in each row highlighted in bold. Note that SVM classifies HSI data, X data, and combined HSI-X data, respectively, while the other approaches focus exclusively on HSI-X data. 

The experimental results show a decreasing trend in accuracy for all comparison methods on the Houston2013 (over 80$\%$), Berlin (around 70$\%$), and DFC2018 (around 60$\%$) datasets. This trend indicates an increasing recognition difficulty across these three datasets, which can be attributed to the differences in data sources, categorization system definitions, and the distribution of training and test samples \cite{my2}. Nevertheless, our CoMiX consistently demonstrated superior performance in terms of OA across all three datasets. Referring to Table~\ref{Houston_acc}, CoMiX achieved the best OA of 95.75$\%$, outperforming SVM, FusAtNet, CALC, Fusion\_HCF, MFT, Fusion-FCN, and Flex-MCFNet by remarkable margins of 31.07$\%$, 6.73$\%$, 9.51$\%$, 6.80$\%$, 7.99$\%$, 11.91$\%$, and 3.99$\%$, respectively. 

Furthermore, the performance of SVM on HSI data significantly outperforms that on X data by a margin of 38.35$\%$, 33.80$\%$, and 14.71$\%$ on Houston2013, Berlin, and DFC2018 data, respectively. This suggests that, as expected, HSIs contain more discriminative features than the X-model. Combining HSI and X data further improves the accuracy of SVM on all three datasets, highlighting the important role of cross-modal feature fusion in vision tasks.

Despite performance improvements when combining HSI and X data, SVM still lags behind other DL-based methods. As a conventional pixel-level classifier, SVM directly processes raw pixels and cannot effectively exploit the discriminative contextual information in the data. The introduction of spatial information with DL-based approaches resulted in consistently improved performance across all three datasets. For instance, the patch-based learning frameworks (i.e., FusAtNet, CALC, Fusion\_HCF, MFT, and Flex-MCFNet) achieved higher OA, exceeding 80$\%$ and 70$\%$ on the Houston2013 and Berlin datasets, respectively, by using attention mechanisms to refine feature representations or perform cross-modality fusion. However, their limited patch size hinders their ability to effectively model long-range spatial dependencies, resulting in the loss of critical details. Overcoming the limited patch size, Fusion-FCN processes larger input images and performs pixel-to-pixel classification. However, its performance is comparable to or even worse than those of patch-based networks. This is because its effective receptive field is still limited by using a stack of $3 \times 3$ convolutions, which prevents Fusion-FCN from modeling long-range dependencies that are essential for segmentation-like tasks.

Compared with the above approaches, our CoMiX takes the lead in both OA and $\kappa $ values. Specifically, CoMiX not only processes larger input images and performs pixel-to-pixel classification, but is also  effective in adaptively modeling local- and long-range dependencies. Additionally, benefiting from the 2D DCN and 3D DCN blocks for adaptive extraction of modality-specific information, as well as CMFeX and FFM for cross-modality feature interaction and fusion, CoMiX results in a robust and comprehensive approach to HSI-X semantic segmentation. Although the AA of CoMiX is slightly lower than that of Flex-MCFNet, FusAtNet, and MFT in Table~\ref{Berlin_acc}, it achieved the best OA and $\kappa $ on all three datasets. Moreover, it also performed well on some challenging classes (i.e., \textit{roads} in the Houston2013 dataset, \textit{industrial area} in the Berlin dataset, and \textit{artificial turf} in the DFC2018 dataset), demonstrating its superiority in handling complex vision tasks with improved accuracy and reliability.

\begin{table*}[htbp]
  \centering
  \caption{Classification accuracy produced by different methods on the Berlin dataset. The numbers within ( ) are the standard deviations of the corresponding metrics. The best result in each row is marked in bold}
   \begin{tabular}{p{4.19em}cccccccccccc}
    \toprule[1pt]
    
    \makecell{Metrics \\ and Class} & \makecell{SVM-X \\ \cite{rSVM}}  & \makecell{SVM-HSI \\ \cite{rSVM}} & \makecell{SVM \\ \cite{rSVM}} &  \makecell{FusAtNet \\ \cite{9150738}} & \makecell{CACL \\ \cite{rCALC}} & \makecell{Fusion\_HCT \\ \cite{rFusionHCT}} & \makecell{MFT \\ \cite{rMFT}} & \makecell{Fusion-FCN \\ \cite{rFusionFCN}}  & \makecell{Flex-MCFNet\\ \cite{10216780}}  & \makecell{CoMiX \\ (Ours)} \\
    \midrule[0.5pt] 
     OA (\%)  & \makecell{27.54  \\ (0)} & \makecell{61.34 \\ (0)} & \makecell{62.98 \\ (0)} & \makecell{73.86\\ (1.84)} & \makecell{75.39 \\ (1.76)} & \makecell{76.46 \\ (1.51)} & \makecell{75.21 \\ (1.59)}  & \makecell{55.14 \\ (1.74)}  & \makecell{70.84 \\ (1.41)} & \makecell{\textbf{76.81} \\ (1.01)} \\
    AA (\%)  & \makecell{20.11 \\ (0)} & \makecell{61.15 \\ (0)} & \makecell{62.37 \\ (0)} & \makecell{66.21\\ (2.08)} & \makecell{63.83 \\ (2.31)} & \makecell{61.47 \\ (2.13)} & \makecell{64.02\\ (2.19)} & \makecell{57.52 \\ (2.32)}  & \makecell{\textbf{68.57} \\ (1.51)} & \makecell{63.31 \\ (0.95)} \\
    $ \kappa\times 100$ & \makecell{10.25 \\ (0)} & \makecell{47.59 \\ (0)} & \makecell{49.53 \\ (0)} & \makecell{60.72 \\ (1.97)} & \makecell{61.89 \\ (2.08)} & \makecell{63.38 \\ (1.92)} & \makecell{63.16 \\ (1.61)} & \makecell{40.23 \\ (1.74)}  & \makecell{58.51 \\ (1.59)} & \makecell{\textbf{63.97} \\ (1.08)} \\
    \midrule[0.5pt]
    C1    & \textbf{88.27} & 80.34  & 80.23  & 50.05  & 47.33  & 59.50  & 79.69  & 43.19  & 74.14 & 65.13  \\
    C2    & 26.18  & 57.41  & 58.34   & 81.09  & \textbf{86.43}  & 85.89  & 79.19  & 54.44  & 70.92 & 84.81 \\
    C3    & 46.43  & 46.43  & 48.30  & 48.51  & 41.13  & 65.57 & 42.84  & 34.15  &53.12 & \textbf{67.20}  \\
    C4    & 0.00  & 76.36  & 84.88  & 85.58 & 80.30  & 85.57  & 87.50  & 69.40  & 83.14 & \textbf{87.86}  \\
    C5    & 0.00  & 74.24  & 74.40  & \textbf{96.38} & 96.27  & 83.54  & 70.64  & 78.15  &77.06 & 79.68  \\
    C6    & 0.00  & 60.85  & 59.75  & 59.74  & 60.14  & 55.03  & 62.27  & 64.16  & \textbf{75.08} & 41.39  \\
    C7    & 0.00  & 27.84  & 27.25  & 29.88  & 26.09  & 12.72  & 32.06 & \textbf{46.77}  & 38.86 & 16.45  \\
    C8    & 0.00  & 65.73  & 65.80  & \textbf{78.42} & 72.95  & 43.92  & 57.98  & 69.92  & 76.28 & 63.93 \\
    \bottomrule[1pt]
    \end{tabular}%
  \label{Berlin_acc}
\end{table*}%

\begin{figure*}[th] 
\centerline{\includegraphics[width=16.5cm]{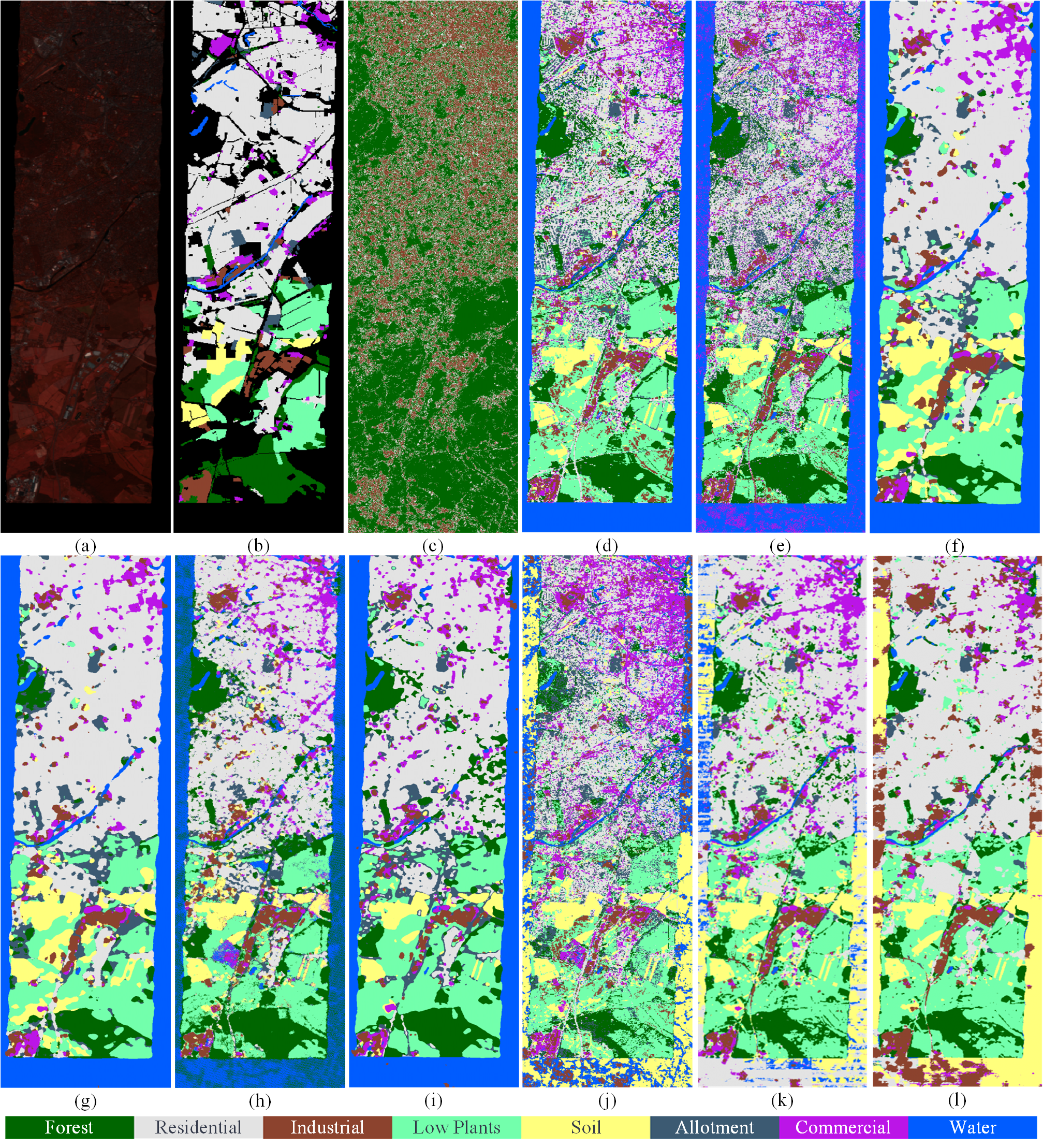}} 
\caption{Classification maps of different methods on the Berlin dataset:  (a) False Color Image, (b) Ground Truth,  (c) SVM-X, (d) SVM-HSI, (e) SVM, (f) FusAtNet, (g) CACL, (h) Fusion\_HCT, (i) MFT, (j) Fusion-FCN, (k) Flex-MCFNet, and (l) CoMiX.}
\label{Berlin_map} 
\end{figure*}

\subsubsection{Qualitative Results and Analysis}
\begin{table*}[htbp]
  \centering
  \caption{Classification accuracy produced by different methods on the DFC2018 dataset. The numbers within ( ) are the standard deviations of the corresponding metrics. The best result in each row is marked in bold}
   \begin{tabular}{p{4.19em}cccccccccccc}
    \toprule[1pt]
    \makecell{Metrics \\ and Class} & \makecell{SVM-X \\ \cite{rSVM}}  & \makecell{SVM-HSI \\ \cite{rSVM}} & \makecell{SVM \\ \cite{rSVM}} &  \makecell{FusAtNet \\ \cite{9150738}} & \makecell{CACL \\ \cite{rCALC}} & \makecell{Fusion\_HCT \\ \cite{rFusionHCT}} & \makecell{MFT \\ \cite{rMFT}} & \makecell{Fusion-FCN \\ \cite{rFusionFCN}}  & \makecell{Flex-MCFNet\\ \cite{10216780}} & \makecell{CoMiX\\ (Ours)} \\
    \midrule[0.5pt] 
    OA (\%)  & \makecell{25.67  \\ (0)} & \makecell{41.15 \\ (0)} & \makecell{56.60 \\ (0)} & \makecell{61.24\\ (0.98)} & \makecell{60.06 \\ (1.56)} & \makecell{62.90\\ (1.18)} & \makecell{60.26 \\ (1.38)}  & \makecell{62.87 \\ (1.47)} & \makecell{63.04 \\ (1.56)}  & \makecell{\textbf{68.26} \\ (1.59)} \\
    AA (\%)  & \makecell{21.93 \\ (0)} & \makecell{36.45 \\ (0)} & \makecell{49.90 \\ (0)} & \makecell{56.17\\ (1.11)} & \makecell{53.26\\ (1.89)} & \makecell{56.84 \\ (2.11)} & \makecell{54.17 \\ (1.79)} & \makecell{57.00\\ (2.39)} & \makecell{57.32 \\ (1.49)} & \makecell{\textbf{63.57}\\ (2.12)} \\
    $ \kappa\times 100$ & \makecell{21.05 \\ (0)} & \makecell{37.50\\ (0)} & \makecell{53.91 \\ (0)}& \makecell{58.79 \\ (1.06)}  & \makecell{57.61\\ (1.50)} & \makecell{60.66 \\ (1.71} & \makecell{57.85 \\ (1.82)} & \makecell{60.62 \\ (1.71)} & \makecell{60.76 \\ (1.62)} & \makecell{\textbf{66.31}\\ (1.21)} \\
    \midrule[0.5pt]
    C1    & 8.11  & 95.82  & 96.18  & 95.27  & 94.60  & 96.19  & 95.47  & 96.53  & 95.54  & \textbf{97.26} \\
    C2    & 90.89  & \textbf{91.86} & 91.59  &87.13  & 89.99 & 86.01  & 90.86 & 87.80  &  86.74 &  85.61 \\
    C3    & 0.00  & 40.50  & 54.81  & 15.72  & 60.15  & 53.30  & 34.23  & 48.79  & 40.42 &  \textbf{84.20}  \\
    C4    & 66.37  & 86.95  & 95.71  & \textbf{97.28}  & 94.43  & 95.39  & 95.50  & 95.99  &  96.61 & 93.29 \\
    C5    & 0.00  & 28.02  & 55.32  & 57.97  & 48.75  & 53.00  & 49.70  & \textbf{60.02}  & 52.86  & 54.95 \\
    C6    & 0.00  & 26.72  & 49.12  & \textbf{85.20}  & 45.86  & 56.58  & 53.93  & 62.17  & 66.74  & 49.13 \\
    C7    & 0.00  & 28.13  & 33.35  & 81.23  & 31.70  & 56.05  & 49.55  & 59.37 &72.85 &  \textbf{86.79}  \\
    C8    & 59.82  & 24.34  & 35.32  & \textbf{79.32 } & 51.96  & 64.65  & 61.78  & 59.33  & 69.66 &  72.84  \\
    C9    & 87.02  & 85.53  & 88.34  & 92.15  & 90.72  & 91.04  & 90.93  & 89.46  &  92.10  & \textbf{95.76} \\
    C10   & 84.09  & 61.28  & 73.35  & \textbf{87.95}  & 65.07  & 56.54  & 81.44  & 78.37  & 85.44 & 82.98  \\
    C11   & 39.57  & 35.86  & 53.98  & 64.73  & 56.84  & 59.94  & 63.31  & 60.62  & 55.15 &  \textbf{77.96} \\
    C12   & 0.00  & 0.00  & 0.00  & 8.61  & 9.36  & 9.02  & 7.11  & 10.05  &  0 & \textbf{14.61} \\
    C13   & 2.53  & 44.05  & \textbf{49.84}  & 33.41  & 49.47  & 44.46  & 40.63  & 45.72  & 41.96  & 49.13 \\
    C14   & 0.00  & 8.52  & 9.45  & 14.22  & 24.81  & 26.75 & 20.67  & 21.65  & \textbf{41.39}   & 19.67  \\
    C15   & 0.00  & 0.05  & 0.47  & 9.43  & 6.94  & 8.76  & 7.28  & 9.04  & 5.10 &  \textbf{33.41} \\
    C16   & 0.00  & 15.56  & 46.19  & \textbf{85.34} & 49.72  & 68.57  & 65.05  & 55.61  & 64.56 &  69.47  \\
    C17   & 0.00  & 0.00  & 0.00  & 0.00  & 0.00  & 0.00  & 0.00  & 0.00  & 0.00  & 0.00  \\
    C18   & 0.03  & 0.83  & 38.39  & 33.78  & 50.35  & 48.36  & 46.16  & \textbf{52.53}  &  38.27  & 47.64 \\
    C19   &  0.16  & 6.74  & 59.47  & 74.12  & 75.12  & 88.12  & 62.39  & 82.19  & 84.65  &  \textbf{96.51} \\
    C20   &  0.01  & 48.22  & 67.13  & 20.48  & 69.41  & \textbf{74.00} & 67.47  & 64.77  & 56.36 & 60.17  \\
    \bottomrule[1pt]
    \end{tabular}%
  \label{DFC_acc}
\end{table*}%

\begin{figure*}[htb] 
\centerline{\includegraphics[width=16.5cm]{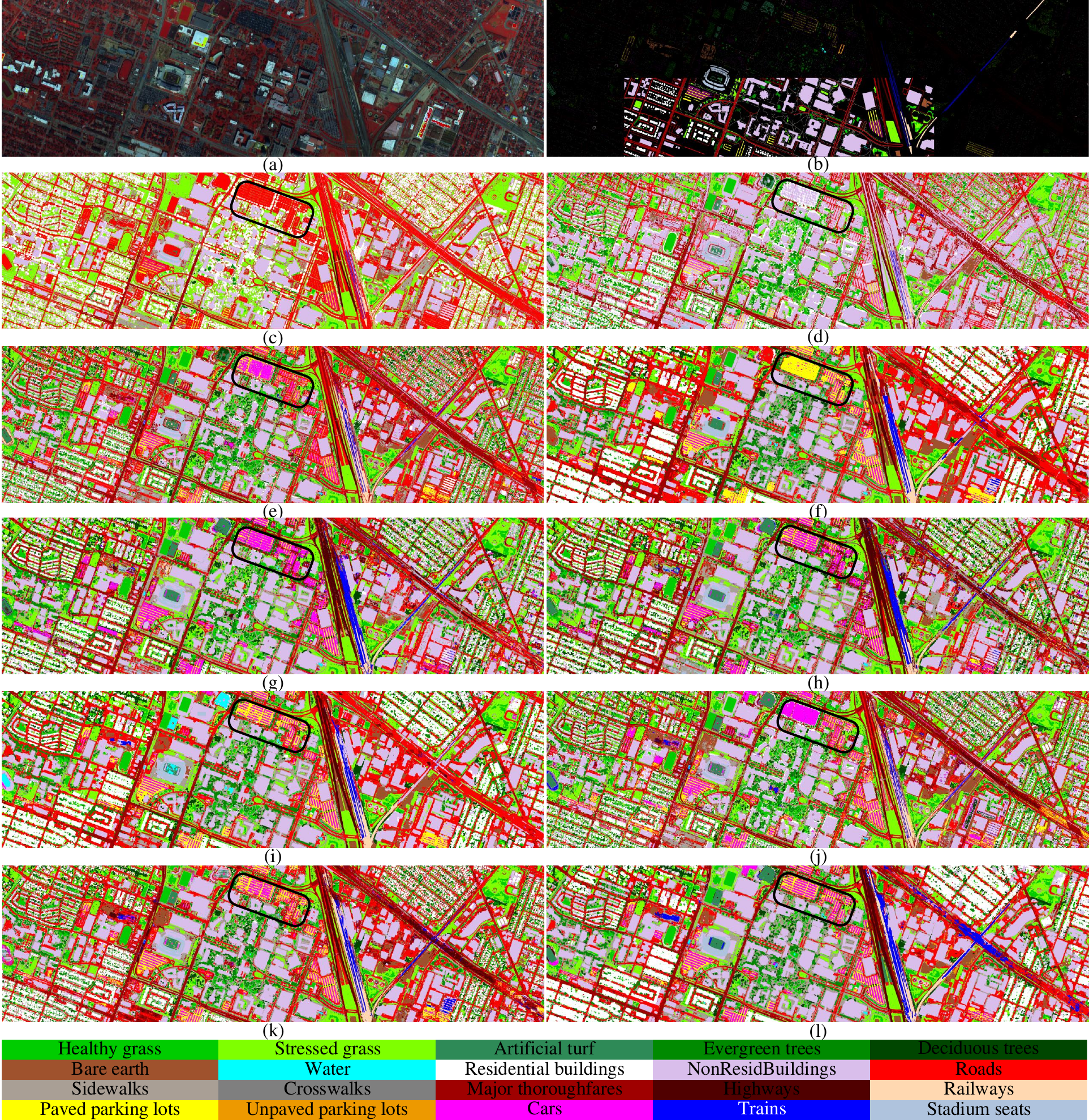}} 
\caption{Classification maps of different methods on the DFC2018 dataset:  (a) False Color Image, (b) Ground Truth, (c) SVM-X, (d) SVM-HSI, (e) SVM, (f) FusAtNet, (g) CACL, (h) Fusion\_HCT, (i) MFT, (j) Fusion-FCN, (k) Flex-MCFNet, and (l) CoMiX.}
\label{DFC_map}
\end{figure*}

The classification maps, along with the corresponding false color and ground truth (GT) images, are shown in Figs.~\ref{Houston_map}–\ref{DFC_map}.

As depicted in Figs. \ref{Houston_map}(c), \ref{Berlin_map}(c), and \ref{DFC_map}(c), SVM can only identify broad categories, such as vegetation, industrial areas, and residential areas, when applied to X data. However, SVM shows improved performance when applied to HSI data, highlighting the importance of rich spectral information for object identification. The fusion of HSI and X data further enhances classification maps, which agrees well with the quantitative results.

In the context of multimodal data fusion and classification, the classification maps produced by SVM exhibit significant salt-and-pepper noise due to the pixel-level fusion strategy. The lack of spatial contextual information poses challenges in distinguishing between overlapping categories, especially land covers in scenarios with similar spectral attributes. 

As a result, patch-based networks like FusAtNet, Fusion\_HCF, and Flex-MCFNet demonstrate improved visual performance by integrating spectral and spatial features and employing adaptive cross-modality feature fusion. However, some algorithms, such as CACL and MFT, can introduce distortions at object boundaries because they assume that each pixel within a patch contributes equally, which  is ineffective in heterogeneous regions. Similarly, Fusion-FCN produced blurred boundary positioning and many errors, especially in low-light or shaded regions. 

In contrast, our CoMiX consistently produced high-quality classification maps with sharper boundaries and smoother objects. In particular, benefiting from the only one 2× downsampling and the fully contextual capturing, our CoMiX  maintains the high-resolution parsing and performs well even under some challenging conditions. For example, the \textit{road}, \textit{residential}, and \textit{green vegetation} in the Houston2013 and DFC2013 scenes, and the well-preserved \textit{low plants} in the Berlin dataset. By enhancing modality-specific information learning and  cross-modality feature fusion, our CoMiX produced highly accurate semantic segmentation maps, particularly excelling in texture and edge details. Furthermore, we selected a region of interest (ROI) (black boxes in Figs. \ref{Houston_map} and \ref{DFC_map}) to highlight differences and facilitate a more intuitive performance evaluation of these methods. As one can see from these ROIs, a notable property of our CoMiX is that it generated maps that show more reliable and finer details. Overall, the qualitative investigation confirms the effectiveness of our CoMiX in various multimodal data fusion scenarios, facilitating robust semantic scene understanding.

\begin{table}[bhtp] 
\caption{Number of Params and FLOPs, Training (Train) and Inference (Infer) time of different methods on the Houston2013 dataset}
\begin{tabular}{p{5.7em} | c c c c}
\toprule[1pt]
Method         & Params     & FLOPs     & Train (s)      & Infer (s) \\ \hline
SVM            & —          & —         & 0.18          & 2.85 \\
FusAtNet       & 36.90M     & 221.61G   & 50583.30      & 57.15 \\
CACL           & 0.34M      & 1.84G     & 1696081351.24 & 54.21 \\
Fusion\_HCF    & 0.43M      & 0.51G     & 2126.89       & 10.45 \\
MFT            & 0.31M      & 1.62G     & 19010.97      & 44.79 \\
Fusion-FCN     & 0.09M     & 6.22G      & 6720.78       & 9.60 \\ 
Flex-MCFNet    & 7.14M     & 83.81G     & 7859.25       & 64.21  \\ \hline
CoMiX          & 21.87M   & 197.02G     & 8282.43       & 7.72 \\
\bottomrule[1pt]
\end{tabular}
\label{TabComplexity}
\end{table}

\section{SENSITIVITY ANALYSIS}
\subsubsection{Efficiency Analysis}
Table~\ref{TabComplexity} compares the number of parameters (Params) and floating point operations (FLOPs), the training time, and the inference time of different methods on the Houston2013 dataset. As observed, SVM takes less training and testing time than the other DL methods. Among these DL methods, CoMiX achieves a significantly faster inference speed despite having higher Params and FLOPs compared to others. As emphasized in previous studies \cite{rDCNv4}, although Params and FLOPs are the commonly used metrics for evaluating model complexity, they do not always accurately reflect model efficiency. In practice, the efficiency of a model is affected by various factors beyond just Params and FLOPs. This observation is also consistent with the patch-based models. For example, among the patch-based models (i.e., FusAtNet, CACL, Fusion\_HCF, MFT, and Flex-MCFNet), Fusion\_HCT exhibits the highest number of Params but the shortest training and inference time.     

The segmentation networks (i.e., Fusion-FCN and CoMiX) demonstrate faster inference speeds than the patch-based networks (i.e., FusAtNet, CACL, Fusion\_HCF, MFT, and Flex-MCFNet), indicating their suitability for real-world applications. This is because segmentation  networks directly produce pixel-level predictions for the entire input image, enabling them to classify the input image in one pass. In contrast, patch-based networks need to generate overlapping patches and then assign labels to the corresponding central pixels to obtain the entire classification map. Consequently, patch-based networks suffer from more redundant computation than segmentation networks, resulting in inefficiency.

Although our CoMiX is inferior to Fusion-FCN in terms of number of Params, FLOPs, training, and inference time, its classification performance significantly outperforms Fusion-FCN. This improvement is attributed to the observation that CoMiX separately uses the 3D DCN and 2D DCN blocks for HSI and X feature extraction, and the CMFeX and FFM modules for cross-modal feature enhancement, interaction, and fusion. In contrast, Fusion-FCN uses only three $3 \times 3$ convolutional layers and three $2 \times 2$ average pooling layers for feature extraction from each modality, followed by a pointwise addition operation for feature fusion. Furthermore, CoMiX implements near real-time inference, demonstrating its potential for real-world applications. In summary, CoMiX exhibits a superior accuracy-efficiency trade-off compared to its counterparts.

\subsection{Ablation Studies}
A series of ablation experiments were conducted on the Houston2013 dataset to verify the importance of different modules: 2D DCN block, 3D DCN block, CMFeX and FFM. The corresponding results are presented in Table~\ref{Ablation}. In our ablation studies, we take a two-branch backbone with ViT blocks as the encoder and the lightweight All-MLP decoder as the baseline. If CMFeX is ablated, the features are extracted independently in their respective branches. If FFM is removed, we simply average the two features for information fusion.

\subsubsection{Effectiveness of the 2D DCN Block}
We use the 2D DCN block to adaptively integrate local- and long-range dependencies in a more computationally and memory efficient way. We first replaced the ViT blocks in the two branches with 2D DCN blocks. The results in Table~\ref{Ablation}(b) show that integrating the 2D DCN block improves the OA by 4.72$\%$ compared with the baseline, demonstrating superior feature representation of the 2D DCN block.

\subsubsection{Effectiveness of the 3D DCN Block}
Next, we employed our 3D DCN blocks to replace the 2D DCN for the HSI branch. As illustrated in Table~\ref{Ablation}(c), after using the 3D DCN block for HSI feature extraction, the OA is improved from 87.23$\%$ to 89.08$\%$, highlighting the superior performance of the 3D DCN block over the 2D DCN block for HSI information learning. This suggests that deploying backbones with 2D DCN blocks and 3D DCN blocks for X and HSI feature extraction, respectively, is a compelling design choice to fully explore the specific features of both modalities.  

\begin{table*}
 \centering
 \caption{Ablation analysis of the proposed CoMiX on the Houston2013 dataset}
\begin{tabular}{p{8em}|ccccccc}
\toprule[1pt]
Method               & (a) Baseline & (b) 2D DCN block & (c) 3D DCN block &(d) CMFeX (spa) & (e) CMFeX (spe)  & (f) CMFeX  & (g) FFM\\ \midrule[0.5pt]
(a) Baseline         & \ding{51} & \ding{51}  & \ding{51}  & \ding{51} & \ding{51} & \ding{51}  & \ding{51}  \\   \midrule[0.5pt]
(b) 2D DCN block    &           & \ding{51}  & \ding{51}  & \ding{51} & \ding{51} & \ding{51}  & \ding{51} \\
(c) 3D DCN block    &           &            & \ding{51}  & \ding{51} & \ding{51} & \ding{51}  & \ding{51} \\
(d) CMFeX (spa) &           &            &            & \ding{51} & \ding{55} & \ding{51}  & \ding{51} \\
(e) CMFeX (spe) &           &            &            & \ding{55} & \ding{51} & \ding{51}  & \ding{51} \\ 
(f) CMFeX            &           &            &            &           &           & \ding{51}  & \ding{51} \\ 
(g) FFM             &           &            &            &           &           &             & \ding{51} \\   \midrule[0.5pt]
OA ($\%$)                & 82.51     & 87.23      & 89.08      & 91.92    & 91.27      & 93.72       &  95.75   \\  
\bottomrule[1pt]   
\end{tabular}
\label{Ablation}
\end{table*}

\subsubsection{Effectiveness of CMFeX} 
We first experimented some variants of CMFeX.  As shown in Table~\ref{Ablation}, “spa" denotes the use of spatial-wise rectification only, and “spe" represents the use of spectral-wise rectification only. It is evident that incorporating either the channel-only or the spatial-only variants of CMFeX results in performance gains. The improvement in OA for the spatial-only and spectral-only variants is 2.84$\%$ and 2.19$\%$, respectively. Replacing either the spatial-only and spectral-only variants with CMFeX leads to further performance gains, confirming the effectiveness of CMFeX for cross-modal feature calibration and enhancement in both spatial and spectral dimensions, which is critical for robust multimodal segmentation. 

\subsubsection{Effectiveness of FFM}
Using CMFeX alone increases the OA to 93.72$\%$, while integrating CMFeX and FFM further improves the OA to 95.75$\%$. This improvement indicates that the combination of CMFeX and FFM modules plays a key role in facilitating HSI-X information fusion. 

By integrating 2D DCN block, 3D DCN block, CMFeX and FFM, our CoMiX achieves improved segmentation results by effectively learning, calibrating, and fusing features from heterogeneous data sources, enhancing its capability for accurate and robust segmentation.

\subsection{Impact of the Number of Training Samples}
\begin{figure}[htbp] 
\centerline{\includegraphics[width=8cm]{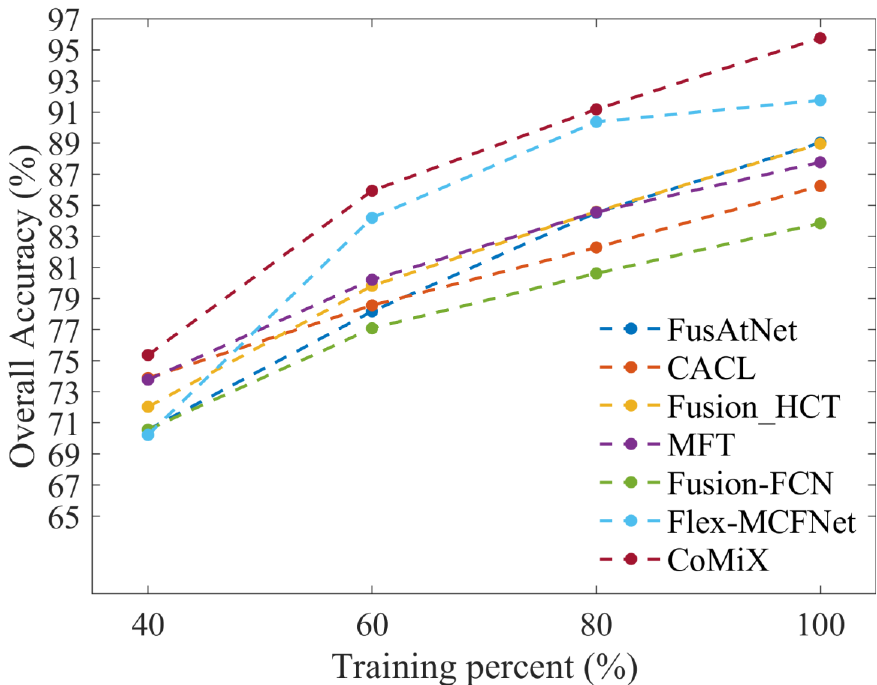}} 
\caption{Classification accuracy of different methods versus the percentage of training samples used on the Houston2013 dataset.}
\label{FigNumTrn} 
\end{figure}

The performance of DL-based methods is strongly influenced by the number of training samples. Therefore, it is necessary to investigate the sensitivity of CoMiX to the number of training samples. In this experiment, we varied the number of training samples per class from 40$\%$ to 100$\%$ with an interval of 20$\%$ on the Houston2013 dataset. The corresponding results are depicted in Fig.~\ref{FigNumTrn}.

Fig.~\ref{FigNumTrn} reveals a rapid decrease in accuracy for all methods as the number of training samples decreases, especially when the percentage of training samples is less than 50$\%$. This highlights the challenge faced by DL-based methods when dealing with limited training data, as insufficient samples often lead to overfitting and poor generalization. However, CoMiX consistently outperforms others across different percentages of training data, demonstrating its outstanding performance and robustness. This superior performance can be attributed to the 2D DCN blocks, 3D DCN blocks, and the CMFeX module of CoMiX, which effectively leverage information from different modalities to enhance feature representation even with fewer samples.

This finding is consistent with our earlier analysis that CoMiX not only boosts performance, but also reduces its dependence on data requirements by replacing the transformer with DCNs. The integration of DCNs allows CoMiX to preserve the inductive bias of convolutions, thereby reducing the model's  dependence on large training datasets. This adaptability is particularly beneficial in scenarios where obtaining a large annotated dataset is challenging, such as in remote sensing and medical imaging, highlighting the potential of CoMiX in real-world applications where data scarcity is a common issue.

In summary, the experiment underscores the importance of robust model design in handling varying amounts of training data, and demonstrates the potential of CoMiX as a reliable method in data-constrained environments.

\section{Conclusion}
In this study, we propose CoMiX, an encoder-decoder cross-modal fusion framework tailored to pixel-wise HSI-X semantic segmentation. In the encoder, 3D DCN blocks and 2D DCN blocks were designed to adaptively extract modality-specific features from HSI and X data, respectively. Additionally, we developed the CMFeX module to recalibrate and highlight modality-specific and modality-shared information while promoting the interaction of cross-modal complementary features across spatial and spectral dimensions. This process ensures that the most relevant features from each modality are emphasized. The enhanced HSI- and X-modality features are then processed by the FFM for cross-modal information fusion.
By integrating 2D DCN blocks, 3D DCN blocks, CMFeX, and FFM into CoMiX, we effectively extract, calibrate, and fuse information from different modalities, creating a robust and efficient framework for multimodal recognition tasks. Experimental results demonstrate that CoMiX achieves leading performance and overcomes the inherent challenges of cross-modal fusion. CoMiX not only ensures comprehensive feature extraction but also facilitates effective interaction between different modalities, thereby improving the accuracy and reliability of segmentation results.

In the future, we will extend CoMiX to handle arbitrary cross-modal fusion scenarios and incorporate three or more types of modalities. This extension will further enhance the flexibility and robustness of our framework, potentially broadening its applicability to a wider range of cross-modal fusion tasks in various fields.

\section*{Acknowledgment}
The authors would like to thank the IEEE GRSS IADF and Hyperspectral Image Analysis Lab at the University of Houston for providing the Houston2013 and DFC2018 datasets.

\ifCLASSOPTIONcaptionsoff
  \newpage
\fi

\bibliographystyle{ieeetr}
\bibliography{ref}


\end{document}